\documentclass{article} 

\usepackage{iclr2027_preprint,times}

\usepackage{hyperref}
\usepackage{url}
\usepackage{xspace}
\usepackage{graphicx}
\usepackage{amsmath}
\usepackage{booktabs}       
\usepackage{dsfont}
\usepackage{mdframed}
\usepackage{subcaption}
\usepackage[T1]{fontenc}
\usepackage{xcolor}
\usepackage[verbatim]{minted}
\usepackage[most]{tcolorbox}
\tcbuselibrary{minted}
\newcommand{\methodname}[0]{\mbox{\textsc{GesBench}}\xspace}

\definecolor{promptbg}{HTML}{F5F4EF}
\definecolor{promptblue}{HTML}{1F5E7A}

\newtcblisting{promptbox}[1]{
  enhanced,
  breakable,
  listing only,
  colback=promptbg,
  colframe=promptbg,
  boxrule=0pt,
  arc=0pt,
  outer arc=0pt,
  borderline west={2.5pt}{0pt}{promptblue},
  left=5mm,
  right=3mm,
  top=1mm,
  bottom=2mm,
  title={#1},
  colbacktitle=promptbg,
  coltitle=promptblue,
  fonttitle=\ttfamily\bfseries\normalsize,
  listing engine=minted,
  minted language=yaml,
  minted style=friendly,
  minted options={
    fontsize=\small,
    breaklines,
    autogobble,
    tabsize=2
  }
}

\title{\methodname: When and Where Do Multimodal Language Models Perceive Social Meaning?}

\author{Zonghuan Li, Hayley Hung, Chirag Raman\\
Department of EEMCS\\
Delft University of Technology\\
Delft, Netherlands \\
\texttt{\{Z.Li-25, H.Hung, C.A.Raman\}@tudelft.nl} \\
\And
}

\iclrpreprintcopy

\begin{document}

\maketitle

\begin{abstract}
Human social perception is incremental: we infer intent before an utterance is complete, revise our interpretation as new cues arrive, and attend to specific facial, gestural, or prosodic signals that shape social meaning. Current multimodal LLM benchmarks largely evaluate final-answer understanding over complete episodes, rewarding post-hoc recognition of social meaning rather than the unfolding perception required for real-time interaction. Drawing on interactionally embedded Gestalt principles of multimodal human communication, we introduce \methodname: a human-grounded benchmark for cue-level social perception in multimodal LLMs. \methodname uses cumulative sub-utterance clips to measure when models form stable interpretations of speaker intention and social affordance, and human gaze-guided corruption to evaluate whether regions attended by humans are functionally important for model judgments. Built from more than $6000$ utterance-level clips from scripted and unscripted conversational datasets, \methodname includes human annotations of unfolding judgments and gaze on a representative subset. Across open and closed multimodal models, we find a clear human-model gap: humans form predictive understandings early and revise more often, while models often hedge or repeat across different clips. \methodname shifts social evaluation from whether models eventually recognize meaning to when and where that meaning is formed. Our codes are publicly available at \url{https://anonymous.4open.science/status/GesBench-share-212C}.
\end{abstract}

\section{Introduction} \label{sec:intro}

Multimodal language models are increasingly evaluated on their ability to understand social meaning: recognizing intentions, emotions, conversational norms, or appropriate responses. Yet, most benchmarks test this ability \textit{after} the model has observed a complete episode. This evaluation setting rewards post-hoc recognition of social meaning, but does not test a capability that is central to real-time interaction: perceiving meaning as the multimodal scene unfolds. 

Human social perception is not delayed until the end of an utterance~\citep{schlangen2011general}. We often form provisional interpretations from partial words~\citep{grosjean1980spoken}, facial expressions, gaze shifts, gestures, pauses, and prosody, and revise these interpretations as new cues arrive. For example, when a friend responds to a vacation suggestion with ``Well, actually I...'', we may already anticipate disagreement, especially if this fragment is accompanied by a furrowed eye brows. If the utterance continues as ``...I was thinking of the exact same place!'', and the expression changes into raised eye brows and a smile, the interpretation is rapidly revised. The social meaning of the utterance is thus not simply recovered from the complete signal but constructed, tested, and updated through \textit{sub-utterance cues} (SUC).

This view is closely aligned with interactionally embedded Gestalt principles of multimodal human communication of \citet{trujillo2023interactionally}: people bind partial multimodal signals into coherent utterance-level interpretations, while also predicting the speaker's intended social action and the response opportunities available to the listener. In other words, social perception is both predictive~\citep{fan2019understanding,ishii2017prediction} and affordance-oriented~\citep{valenti1991social}: perceivers do not only infer what the speaker means, but also what actions the situation makes possible. 

For embodied agents and multimodal LLMs that are expected to participate in face-to-face interactions, this distinction matters~\citep{fung2025embodied, zhang2025mindpower}. An agent that can answer correctly after seeing a complete video may still fail to support fluent interaction if it cannot form useful provisional interpretations, attend to communicatively relevant cues~\citep{qi2025capturing,upadhyay2026time}, or update its beliefs smoothly as the utterance progresses. Existing multimodal and social benchmarks evaluate many important abilities, including video understanding~\citep{zhang2025mmla,li2024mvbench}, social reasoning~\citep{mathur2025social}, emotion recognition~\citep{lian2024affectgpt}, and social acting~\citep{wu2025social}. However, they are less suited to measuring the unfolding process by which social meaning is formed from fine-grained multimodal cues.

We introduce \methodname, a human-grounded benchmark for evaluating cue-level social perception in MLLMs. \methodname operationalizes interactionally embedded Gestalt perception along two complementary axes. First, to measure temporal unfolding, we decompose each utterance into cumulative  clips and ask both humans and models to infer the speaker's intended social action and the listener's possible response actions. This allows us to measure when are perceptions formulated and how they update through the process. Second, to measure spatial grounding, we collect human eye gaze while annotators interpret utterances, construct human-attended cue profiles, and corrupt relative regions to test whether those cues are functionally important for models.

Our contributions are three-fold: \textbf{(1)} We formulate cue-level social perception as an evaluation problem for MLLMs, shifting the focus from final-answer recognition over complete videos to the unfolding process by which social meaning is formed. \textbf{(2)} We propose a human-grounded spatio-temporal evaluation protocol: cumulative sub-utterance clips measure how interpretations update, while gaze-guided corruption tests whether perturbing human-attended regions changes model judgments.
\textbf{(3)} We evaluate open and closed source MLLMs and identify a human model gap in unfolding social perception: humans make earlier commitment in predicting, revise more often while models tend to remain uncertain with little information and make fewer revisions.

Together, these results show that current multimodal LLMs can often recover social meaning after observing an utterance, but still differ from humans in how that meaning is formed. \methodname provides a benchmark for studying this gap and for developing models that are more predictive, interactive, and attentive to the fine-grained cues that structure real-time communication.

\begin{figure}[t]
  \centering
  \includegraphics[width=0.9\textwidth]{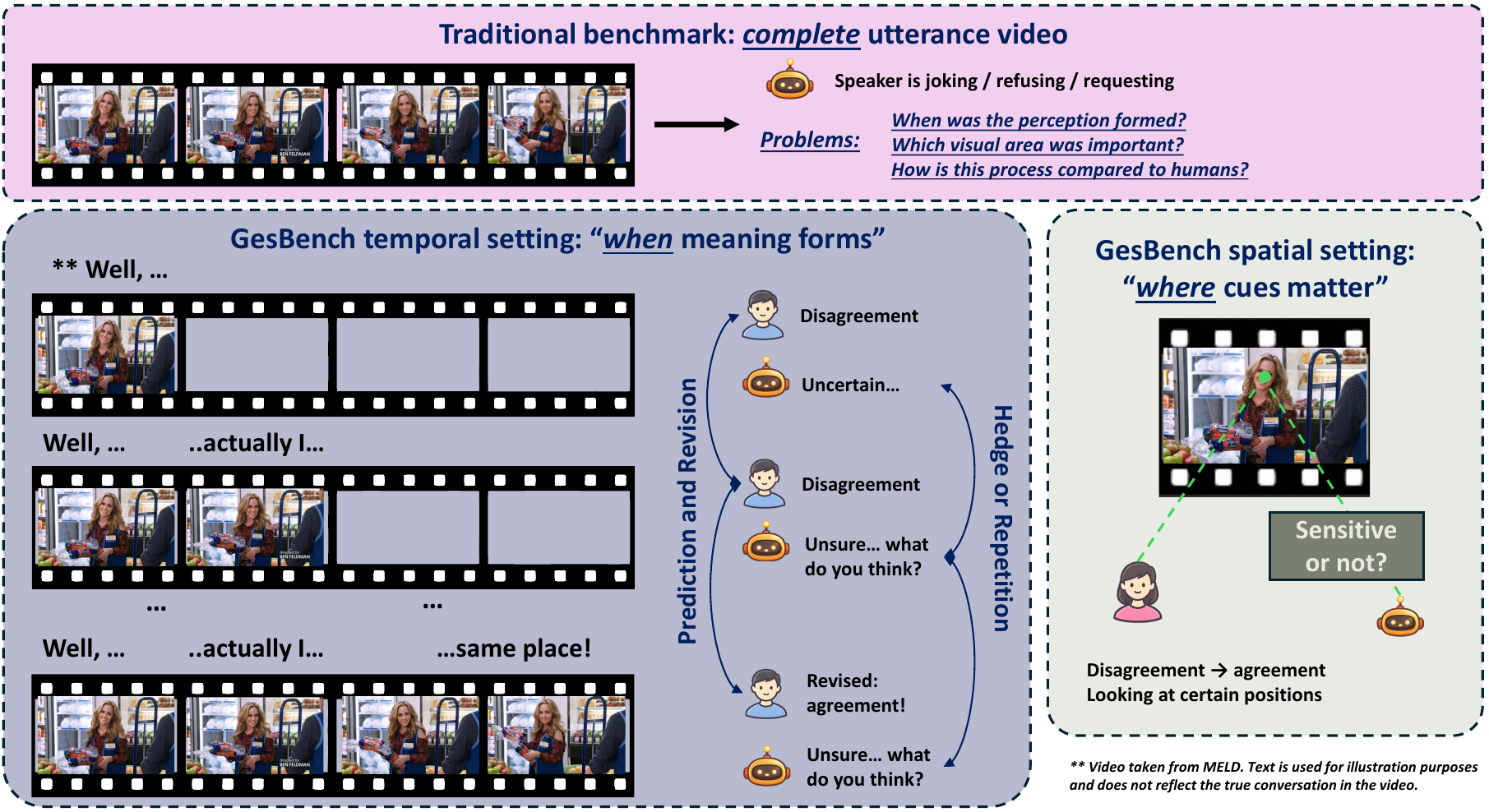}
  \caption{Instead of only asking model perception after the complete utterance, \methodname measures \textbf{when} and \textbf{how} interpretations update. It also collects human gaze during utterance interpretation to compare model judgments of sensitivity to human-attended visual cues. Together, these temporal and spatial probes evaluate when, where, and how social meaning is formed.}
  ~\label{fig:teaser}
  \vspace{-1em}
\end{figure}

\section{Related Works}


\paragraph{MLLM benchmarks.} An abundance of benchmarks have been proposed to measure multi-modal perception of MLLMs. ~\citet{liu2024mmbenchmultimodalmodelallaround, meng2024mmiu} proposes comprehensive image understanding evaluations for MLLMs, including general perception and reasoning tasks. There are also benchmarks for specializing domains, for example on object hallucination~\citep{li2023evaluating}, identifying truthful statements~\citep{kang2025can}, and moral reasoning~\citep{jiao2025llm}. Specifically, for video understanding, ~\citet{fu2025video, li2024mvbench} provide general evaluation frameworks regarding videos from film clips to everyday recordings. Other benchmarks focus on social abilities of MLLMs. ~\citet{zhang2025mmla} benchmarks MLLMs in perceiving multi-modal communication signals with a distribution of social tasks. ~\citet{strachan2024testing} has tested the theory of mind (ToM) in MLLMs by testing their abilities in interpreting the other party's minds through tasks such as identifying indirect requests. ~\citet{mathur2025social} benchmarks MLLMs in understanding long videos and through correct reasoning traces, but it does not explicitly address the temporal updating characteristic. For multi-agent tasks, ~\citet{zhou2023sotopia, wu2025social} evaluates the performances of MLLM-powered agents under certain social interaction settings and against various social targets.

\paragraph{Gestalt evaluations in AI models.}Attempts have been made to study gestalt principles on both traditional machine learning tasks like computer vision as well as modern MLLMs. ~\citet{sha2016gestalt} has studied classification problems with regards to the gestalt principles in human vision to evaluate whether the models are recognizing holistic objects. ~\citet{kim2021neural} also points out that some computer vision models are exhibiting gestalt principles in terms of closure recognition. Regarding MLLMs, ~\citet{dayanandan2024dual} has leveraged the dual thinking framework~\citep{kahneman2011thinking} to help improve MLLMs' performances when facing adversarial data. However, these benchmarks either focus on low-level Gestalt principles, or do not specifically address Gestalt perception in social interaction. \methodname, to our knowledge, is the first benchmark built with real social data and designed to incorporate gestalt features in model and human perceptions.

\paragraph{Spatial grounding and visual cue evaluation in models.}
Prior work studies whether vision-language models ground answers in appropriate image regions using heatmaps~\citep{selvaraju2016grad, christian2026attention}, attention analyses~\citep{abnar-zuidema-2020-quantifying}, or region-level grounding objectives~\citep{zhang2025gpt4roiinstructiontuninglarge,chen2023shikra,yuan_osprey_2024,qi2025capturing}. These methods provide useful diagnostics, but they are typically post-hoc or centered on static image understanding rather than the unfolding perception of social meaning in video. Recently more works have been proposed to study whether aligning with human eye gaze help model perception~\citep{peng2025eye,pani2025gaze}. \methodname instead uses human gaze collected during utterance interpretation to construct human-attended profiles, and evaluates whether corrupting these regions changes model judgments more than corrupting matched comparison regions.

\begin{figure}[ht]
  \centering
  \includegraphics[width=0.85\textwidth]{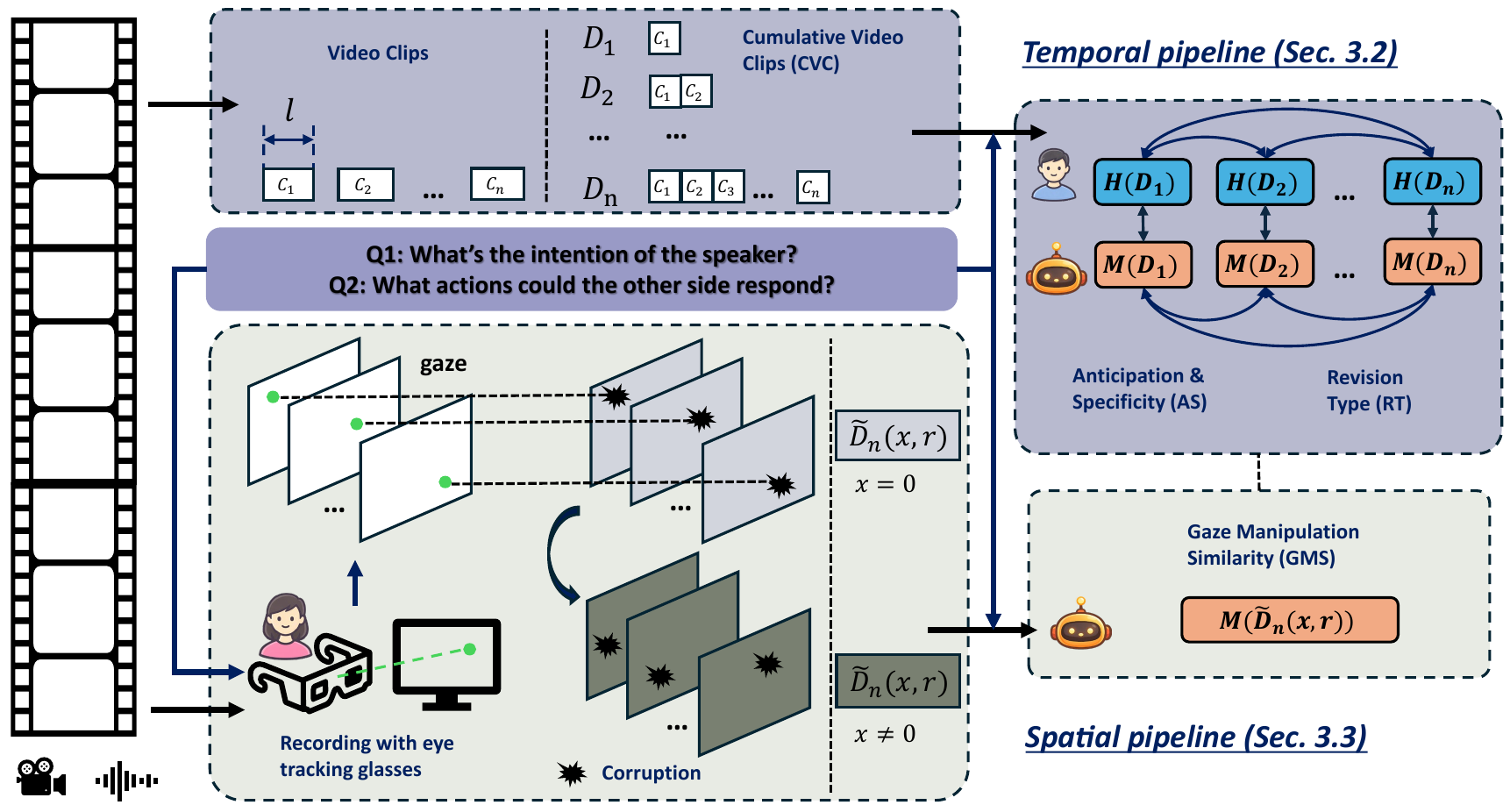}
  \caption{Illustration of the evaluation pipeline. The temporal pipeline takes video inputs, cut into small clips, and build CVC which is then given to both human and models. The spatial pipeline measures human eye gaze data, builds a gaze profile, and does corruption to make gaze based corruption videos with varying radius and offset to prompt the models. Finally, we compute specificity, anticipation, revision type, and gaze manipulation similarity for metrics.}
  ~\label{fig:pipeline}
\end{figure}

\section{The \methodname Benchmark} \label{sec: method}

\methodname combines scripted and unscripted conversational data to cover both clearly bounded utterances and natural dyadic interaction. Scripted clips are from MELD~\citep{poria2019meld} and MIntRec2~\citep{zhang2024mintrec2} while Seamless Interaction~\citep{agrawal2025seamless} provides unscripted face-to-face conversations. Our benchmark contains more than $6000$ utterance-level clips, with human text and gaze annotations collected on representative subsets with around $600$ videos. Additional dataset information and selection rationale are provided in Appendix~\ref{sec:dataset_info}. The overarching structure of \methodname is shown in figure~\ref{fig:pipeline}. In \ref{subsec:task_designing} we illustrate our design of tasks, prompts, and questions. Sections~\ref{subsec:temporal_benchmark} and \ref{subsec:spatial_benchmark} introduces temporal/spatial pipelines. In~\ref{subsec:metrics} we demonstrate how metrics are defined and calculated.

\subsection{Prompts and questions} \label{subsec:task_designing}

We leverage single utterances as the basic unit of perception. Referring to the Gestalt Framework framework~\citep{trujillo2023interactionally,holler2019multimodal} and 
the social affordance theory~\citep{valenti1991social}, we ask two questions to both human and models, with the full prompts in Appendix~\ref{subsec:model_prompting}:

\begin{promptbox}{GESBENCH PROMPT (simplified)}
Q1: What is the intention of the speaker? 
Q2: What actions could the other side take in response?
\end{promptbox}


\subsection{Temporal pipeline} \label{subsec:temporal_benchmark}
Inspired from approaches which provide models with incremental language~\citep{madureira2020incremental, kahardipraja2021towards} we create cumulative versions of each utterance video. Each cumulative clip contains the signal observed from the start of the utterance up to a given point and are given to the models incrementally context. Concretely, we start with a raw video $V$ of single utterance. To create cumulative short clips, we first cut it into short clips of equal length $l$: $C_1, C_2 \dots, C_n$ and let $\text{Concat}[C_1, C_2, \dots, C_n]$ to be their sequential concatenation. We define the set of \textbf{Cumulative Video Clips (CVC)} with respect to $V$, by concatenating the clips cumulatively:
\begin{equation}
    D_k = \text{Concat}[C_1, \dots, C_k], \quad k=1,2,\dots, n
\end{equation}
By design, each $D_k$ has length $kl$ (we take $l=0.5$~s in practice). The CVC set is then given to both human annotators and the models. We obtain an answer for each video and for both $Q_1$ and $Q_2$. We denote human response for a video $D_k$ under question $Q$ to be $H_Q(D_k)$, and model response to be $M_Q(D_k)$. In the temporal benchmarking pipeline, this would yield the human response set $\{H_Q(D_k)\}_{k=1}^n$ and model response set $\{M_Q(D_k)\}_{k=1}^n$, where $Q\in \{Q_1, Q_2\}$. 

To obtain the human answers for CVC, we employ a crowdsourcing approach. Paid annotators are recruited on Prolific~\footnote{https://app.prolific.com/} platform, and the annotation is performed with Covfee~\citep{quiros2022covfee}. We ask them to speak out the answers, which are then transcribed using WhisperX~\citep{bain2022whisperx}. The answers are then extracted per question. A model reply answers both questions in one go and we split it into $Q_1$ and $Q_2$ with an LLM instruction to extract the answers verbatim; Appendix~\ref{subsec:extraction_fidelity} to~\ref{subsec:early_profile} shows that our conclusions do not depend on this step. Annotators already answer each question in its own field. Sometimes an annotator says ``nothing has changed'' or similar even when we prompted them to provide reasons. To mitigate this, we carry their previous answer forward when this happens. Appendix~\ref{subsec:human_annotation} and~\ref{sec:experimental_setup} details the annotation setup, model prompts, and postprocessing.

\subsection{Spatial pipeline} \label{subsec:spatial_benchmark}
\begin{figure}[ht]
  \centering
  \includegraphics[width=0.85\textwidth]{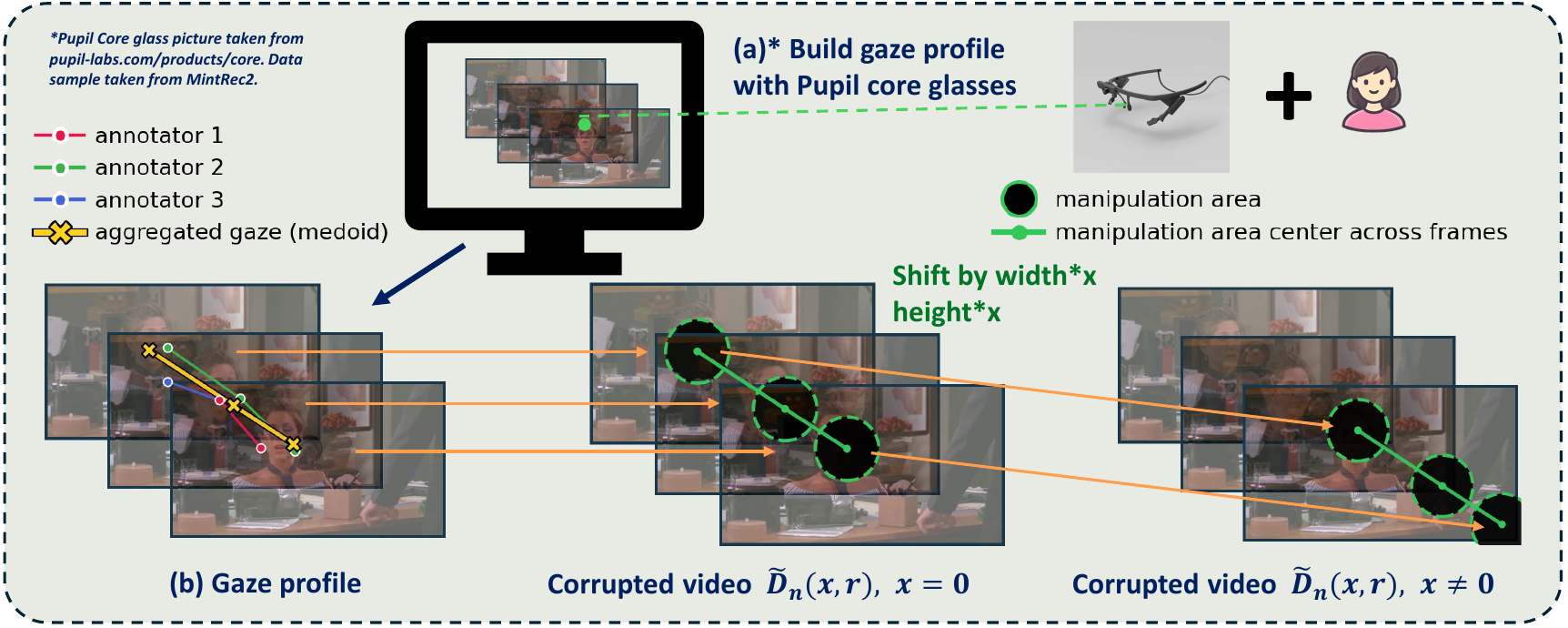}
  \caption{Process of building human attention profile. (a) We record participants' eye gaze data with Pupil Core glasses while watching videos; (b) eye gaze profile by aggregation. Each annotator's gaze for a frame is shown. (c) Build per-frame corrupted video $\widetilde{D}_n(x,r)$ with $x=0$, by blocking the region centered at the gaze point. (d) Build $\widetilde{D}_n(x,r)$ with $x\neq 0$, shifting the manipulation area. A complete set of video manipulation sample is provided in the supplementary material.}
  ~\label{fig:gaze_pipeline}
  \vspace{-1.5em}
\end{figure}
In the spatial domain, interactionally embedded Gestalt perception is reflected in how perceivers bind communicatively relevant visual signals while filtering out incidental ones. Human viewers do not attend uniformly across a scene; they often orient toward regions that afford social interpretation, such as faces, gaze direction, hands, body orientation, and visible articulators. This motivates us to measure whether certain regions attended by humans are important to model perception or not. 

To obtain a human-grounded cue profile for a given clip, we collect eye gaze while annotators watch complete utterance videos and answer the two benchmark questions. For each video $V$, annotators wear eye-tracking glasses while viewing. The recorded gaze points are mapped back to the video, producing a per-frame profile of human attention(Figure~\ref{fig:gaze_pipeline}). 
On top of this, we construct a series of corrupted videos. For a displacement $x$ and a radius $r$, let $\widetilde{D}_n(x,r)$ be the video obtained from $V$ by drawing on every frame a circular mask of radius $r$ times the shorter edge of the video. Its center is determined by $x$: $x=0$ means centered at human gaze point, and non-zero means shifting the disc by an offset of $xw$ horizontally and $xh$ from the gaze center: $w,h$ means width and height of the video respectively. This way, we obtain a grid of varying $(x,r)$s and a control of untouched video with $r=0$. In practice we choose $x \in \{0, 0.1, 0.3, 0.5\}$ and $r \in \{0.1, 0.2, 0.3\}$. The corrupted videos are then given to the models with the same prompts, and the answers are extracted as described in~\ref{subsec:anwer_processing}.


\subsection{Metrics} \label{subsec:metrics}
%
Most of our analysis are based on the semantic similarity of two texts $T_1$ and $T_2$. A pre-trained language encoder $\mathcal{E}$ is used to map responses into an embedding space. We employ the cosine similarity: $S(T_1,T_2)=
\langle \mathcal{E}(T_1),\mathcal{E}(T_2)\rangle
/(\|\mathcal{E}(T_1)\|\cdot \|\mathcal{E}(T_2)\|)$,where $\langle\cdot,\cdot\rangle$ denotes the inner product. We use all-MiniLM-L6-v2~\footnote{https://huggingface.co/sentence-transformers/all-MiniLM-L6-v2} as the text encoder. Following~\citet{trujillo2023interactionally,holler2019multimodal}, we measure 1) whether the annotator makes early and plausible answers to the entire utterance; 2) whether the annotator revises their answers in the process; and 3) whether human attended cues are important to models with following metrics:


\paragraph{Specificity.} We first measure if an author (model/human) is actually providing perceptions specific to the current video $V$, a failure known as generic responses in dialogue~\citep{li2016diversity}. We employ a similar measure as~\citet{wang2020compare} to distinguish model answers from generic ones: how far an annotation is distinctive compared to other answers from the same annotator. Specifically, for video $V$ and $V'$ with different clip numbers $n$ and $n'$, we define $\kappa(k,V,V') = \mathrm{round}(kn'/n)$ to be the clip of $V'$ at the progress in the utterance as clip $k$ of $V$. Following the context in~\ref{subsec:temporal_benchmark}, we use $D_k(V)$ below to denote the $k$-th cumulative clip of video $V$. We measure this on the subset with at least 2 annotators $\mathcal{T}_2$ as mentioned in Appendix~\ref{subsec:dataset_info}:
\begin{equation}
  \mathrm{Spec}_Q(A,V,k) = 1 - \frac{1}{|\mathcal{T}_2\setminus\{V\}|}\sum_{ V'\in \mathcal{T}_2\setminus\{V\}} S\!\left(A_Q\big(D_k(V)\big),\, A_Q\big(D^{(V')}_{\kappa(k,V,V')}\big)\right).
\end{equation}
where $A=H$ or $M$ means either model or human. This comparison is done per perceiver, by averaging over videos $V$ and clip positions $k$ into the final $\mathrm{Spec}_Q(A)$. This score is low for perceivers whose answers are very similar across all videos at similar progress positions, and is high for perceivers who write specific perceptions for each video. Note that natural similarities among videos may influence its value, so we compare on a relative instead of absolute scale for $\mathrm{Spec}$.

\paragraph{Anticipation.}

We also measure if the model's early answers are similar to human final perception. We employ a ranking approach, aiming to measure how whether an perceiver's partial perception $A_Q(D_k(V))$ is more similar to \textit{this} video's final human perception $H(V)$, than \textit{other} videos' final human perceptions $\{H^a_Q(V')\}_{V' \neq V}$. Specifically, 
\begin{equation}
  \mathrm{Ant}_Q(A,V,k) = \frac{1}{|\mathcal{A}_V|}\sum_{a \in \mathcal{A}_V} \mathrm{R}\Big(A_Q\big(D_k(V)\big);\ H^a_Q(V),\ \big\{H^a_Q(V')\big\}_{V' \neq V}\Big).
\end{equation}
where $\mathcal{A}_V$ is the annotator set (for humans, this means other annotators), and $\mathrm{R}$ is the ranking measure similarly employed in~\cite{hanley1982meaning}:
\begin{equation} \label{eq:rank}
  \mathrm{R}\big(q;\, t,\, \mathcal{D}\big) = \frac{1}{|\mathcal{D}|}\sum_{d \in \mathcal{D}} \left( \mathds{1}\big[S(q,t) > S(q,d)\big] + \frac{1}{2}\,\mathds{1}\big[S(q,t) = S(q,d)\big] \right)
\end{equation}
By this design, $\mathrm{Ant}$ measures whether models are aligning with annotator answers, and for humans this provides a comparable measure on agreement over annotators. Both $\mathrm{Spec}$ and $\mathrm{Ant}$ are averaged over the first third of each utterance: $k/n \le 1/3$.




\paragraph{Revision Types (RT).} Gestalt perception also includes actively revising one's own answers. A direct semantic comparison, however, is insufficient to tell the differences, since modifications such as adding a clause, restating a reading, and completely throwing out the previous reading can change an answer similarly in embedding space. To this end, also inspired from the Similarity-Num Steps measure in Social Genome~\citep{mathur2025social}, we consider the natural-logic relations~\citep{maccartney2009extended} with the entailment between adjacent clips. Specifically, we classify each pair of consecutive answers $\big(A_Q(D_{k-1}), A_Q(D_k)\big)$ with a NLI cross-encoder~\citep{liu2019roberta} which discerns the entailment between two texts in both directions. We use $\rho$ for the word-sequence ratio between the two answers, and consider the relationship classes between adjacent responses:
\begin{center}
\footnotesize
\begin{tabular}{ll}
  \textbf{copy} & $\rho \geq \tau_c$: the same content; not sent to the encoder. \\
  \textbf{restate} & the same reading in other words. \\
  \textbf{elaborate} & the reading is kept and something is added. \\
  \textbf{generalize} & the reading is kept and something is dropped or hedged. \\
  \textbf{contradict} & contradiction in either direction: the reading is reversed. \\
  \textbf{shift} & neither entailment nor contradiction: a different reading. \\
\end{tabular}
\end{center}
In practice we choose $\tau_c=0.9$. Details of this classification is provided in Appendix~\ref{subsec:revision_details}. The first four classes keep the previous reading and the last two replace it, so we call \textbf{contradict} and \textbf{shift} together a \emph{reading change}. A perceiver that actively reorganizes its interpretation produces many reading changes; one that repeats itself produces many copies.

\paragraph{Gaze Manipulation Similarity (GMS).}
To compare the masked video perception to original ones, we do the cosine similarity comparison between masked and unmasked videos:
\begin{equation} \label{eq:gsim}
  \mathrm{GMS}_Q(\widetilde{D}_n(x,r)) = S\!\left(M_Q\big(\widetilde{D}_n(x,r)\big),\ M_Q\big(\widetilde D_n(0,0)\big)\right)
\end{equation}
We run with varying $(x,r)$, and also experimented with and without the audio contents. Model responses $M_Q$ are obtained over every masked/unmasked videos and GMS is computed by averaging.

\paragraph{Terminological caution.}
We intentionally avoid referring to GMS as saliency measures. This distinction is important because the term ``saliency'' has been used inconsistently across the literature for heatmaps, salient maps, etc.~\cite[Sec.~5.3]{arrieta2020explainable}. Just as~\citet{raman2023did} argued in their work about information saliency, perturbation-based methods can create semantically invalid inputs~\citep{adebayo2018sanity,lapuschkin2019unmasking}~\cite[Sec. 9.5.3.3]{molnar2020interpretable}, making it unclear whether changes in model output reflect the importance of the removed cue or artifacts introduced by the intervention. Our analysis is designed with these limitations in mind. We build an empirical profile of where human observers look while interpreting multimodal utterances; it is not a ground-truth map of perceptual or causal significance. Likewise, GMS measure how much a model's open-ended judgment changes when human-attended or comparison regions are corrupted. They therefore quantify intervention-based corruption sensitivity, not preattentive or causal necessity. A low GMS indicates that the model is sensitive to corrupting human-attended regions; it should not be interpreted as showing that those regions are intrinsically salient for the model.

\section{Experiments} \label{sec:experiments}

We only consider MLLMs which handle both video and audio inputs for Gestalt principles. We choose \textbf{Qwen3-Omni-30B~\citep{Qwen3-Omni}, Qwen2.5-Omni-7B/3B}~\citep{xu2025qwen25omnitechnicalreport}, \textbf{Ming-Lite-Omni}~\citep{ai2025ming}, and \textbf{Gemma 4-E4B}~\citep{gemmateam2026gemma4technicalreport} as open-source models, and \textbf{Gemini2.5/3.6-flash}~\citep{comanici2025gemini} as closed-source models. We run open-source models locally and the closed-source models through API.

\subsection{Temporal results: specificity, anticipation, and revision} \label{subsec:commitment_results}

Temporal results are done with text annotated videos. $\mathrm{Spec}$ and $\mathrm{Ant}$ (Appendix~\ref{subsec:dataset_info}) with 95\% bootstrap intervals over videos. The $\mathrm{Spec}$, $\mathrm{Ant}$, and RT of Section~\ref{subsec:metrics} separate humans from models both in how specific they perceive and how they revise.

\begin{figure}[ht]
  \centering
  \begin{subfigure}{0.4\textwidth}
      \includegraphics[width=\linewidth]{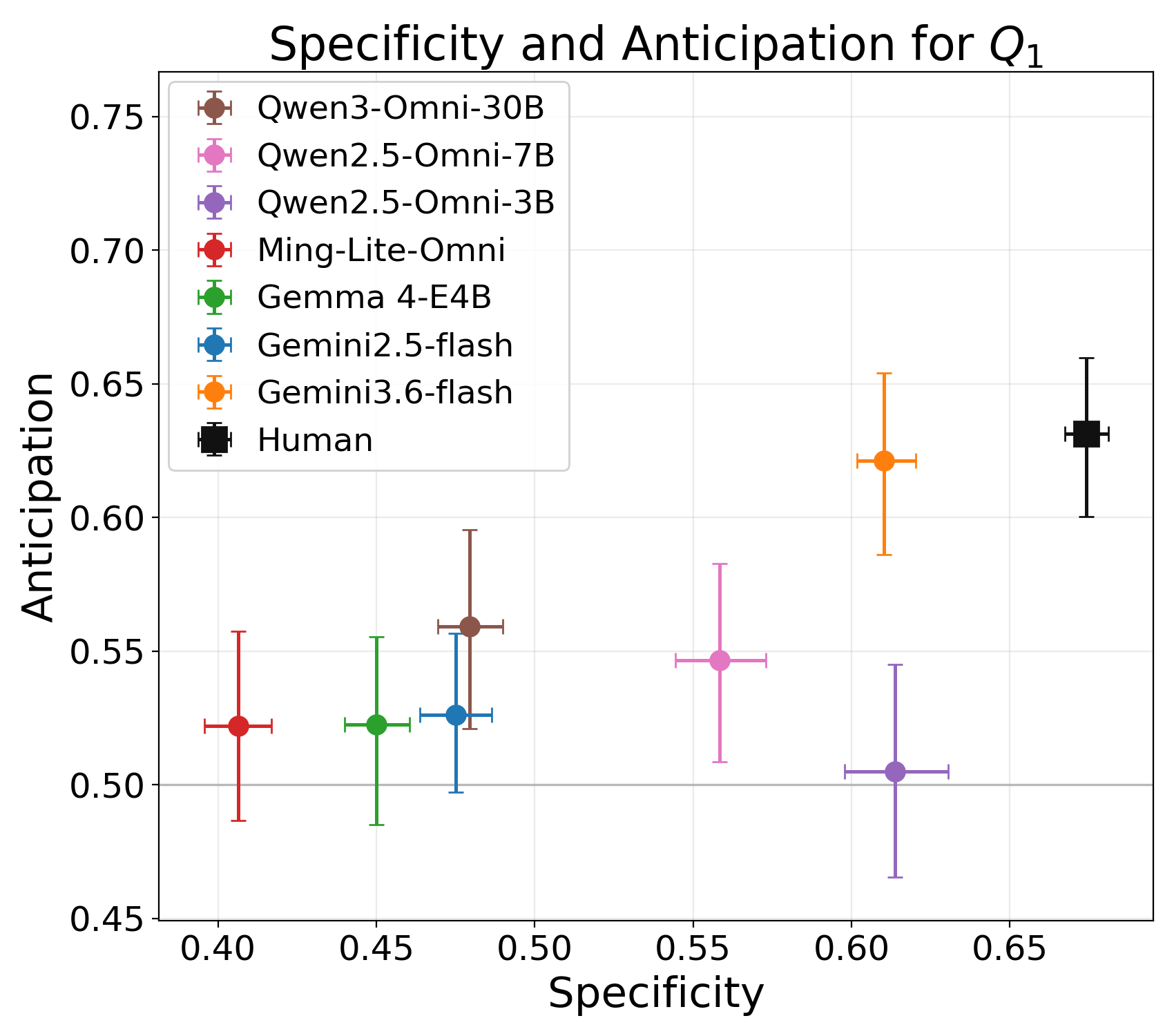}
      \label{fig:specant_q1}
  \end{subfigure}
  \begin{subfigure}{0.4\textwidth}
      \includegraphics[width=\linewidth]{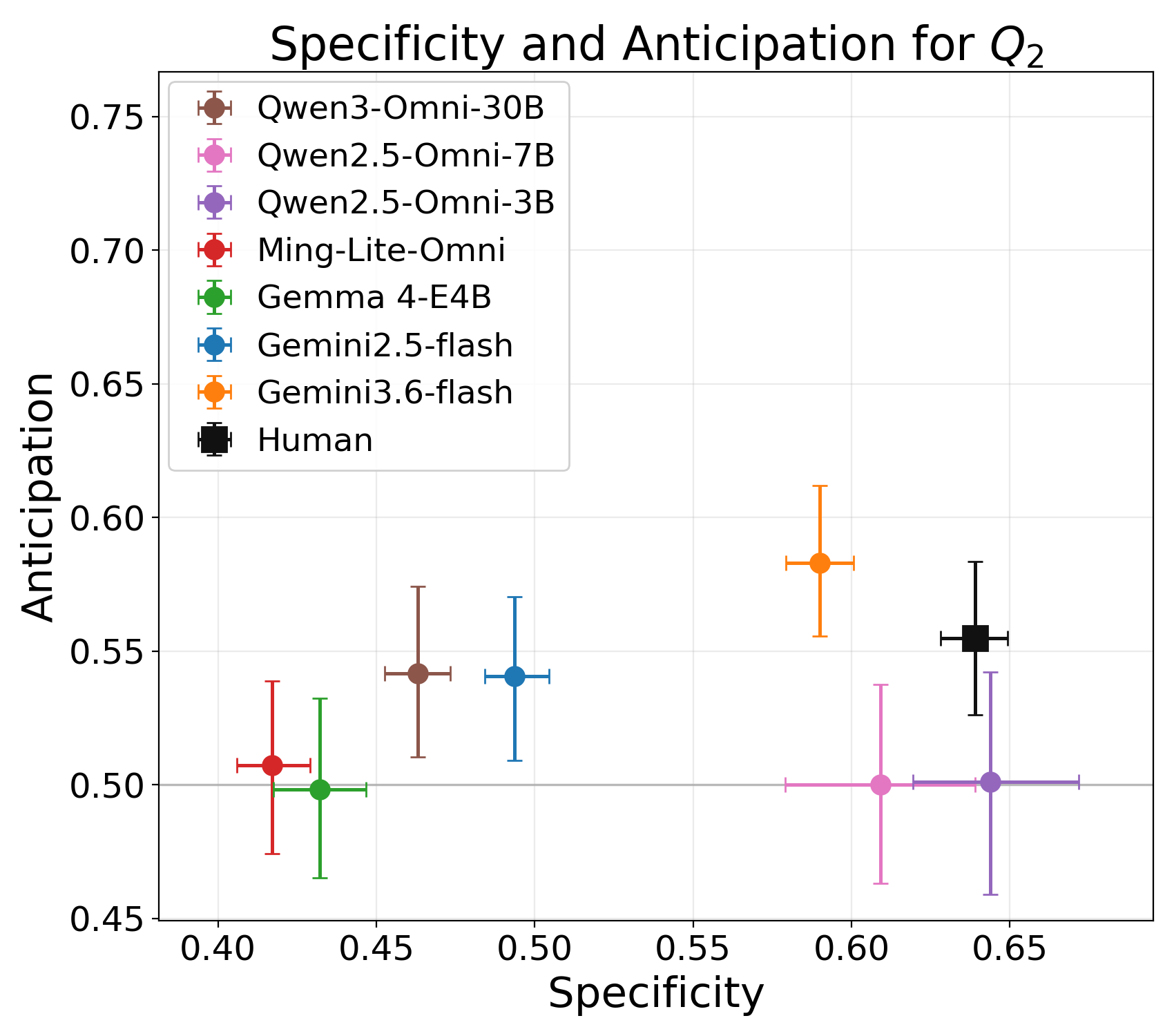}
      \label{fig:specant_q2}
  \end{subfigure}
  \caption{Specificity vs.\ anticipation over the first third of sub-utterance answers for each video, for $Q_1$ (left) and $Q_2$ (right). Anticipation is $0.5$ at chance. Bars are 95\% bootstrap intervals over videos.}
  \label{fig:specant}
\end{figure}

\paragraph{Humans commit early and align more with others's final perception.}
Figure~\ref{fig:specant} places each perceiver's early answers (first one third of answers per video) in the $(\mathrm{Spec}, \mathrm{Ant})$ plane. Humans sit in the top-right corner on both questions, especially for $Q_1$. In the first third of an utterance, their readings are already specific to it and agree with where other annotators end up. This aligns wtih Gestalt account: a single coherent reading on incomplete signal. Models fail to reach human performance in general for both questions.

For $Q_1$, the distinction is clear: all models are not as committed as human, and many models' $\mathrm{Ant}$ are including chance level such as Gemma and Ming-Omni-Lite, showing that they are neither specific enough to the current video nor reaching to another human perception. Gemini3.6-flash is closest to human but still has a gap of around $0.07$ in $\mathrm{Spec}$. For $Q_2$, the gap between human and models are smaller. Gemini3.6-flash surpasses human at around $0.03$ in $\mathrm{Ant}$ but at a cost of $0.05$ inferiority in $\mathrm{Spec}$. Qwen2.5-Omni-7B/3B, on the other hand, are reaching human level $\mathrm{Spec}$, but with $\mathrm{Ant}$ at chance level. This shows that in general human annotators make predictions early and what they perceive is already in more agreement with other humans, compared to models.


\begin{figure}[ht]
  \centering
  \begin{subfigure}{0.49\textwidth}
      \includegraphics[width=\linewidth]{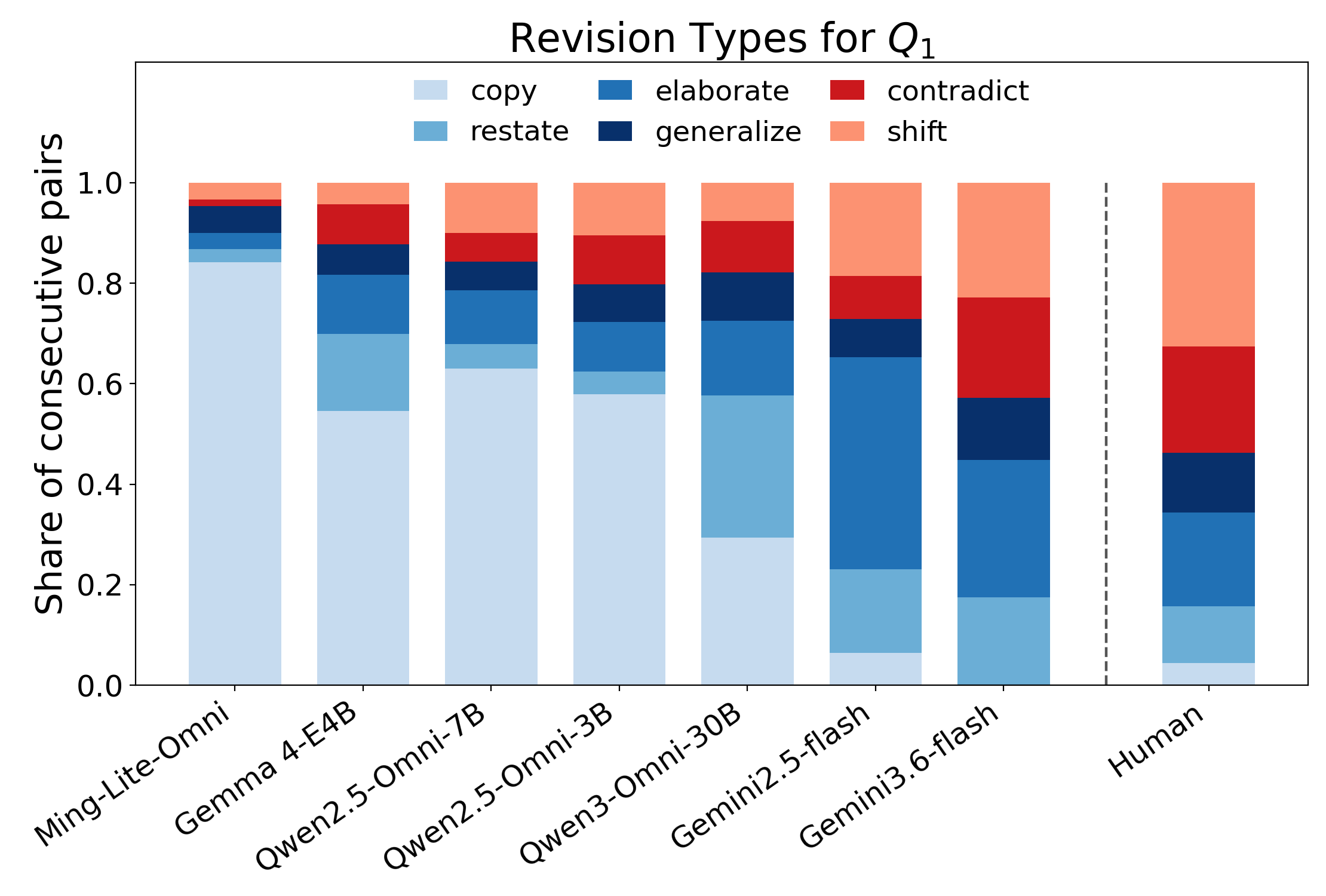}
      \label{fig:revision_q1}
  \end{subfigure}
  \begin{subfigure}{0.49\textwidth}
      \includegraphics[width=\linewidth]{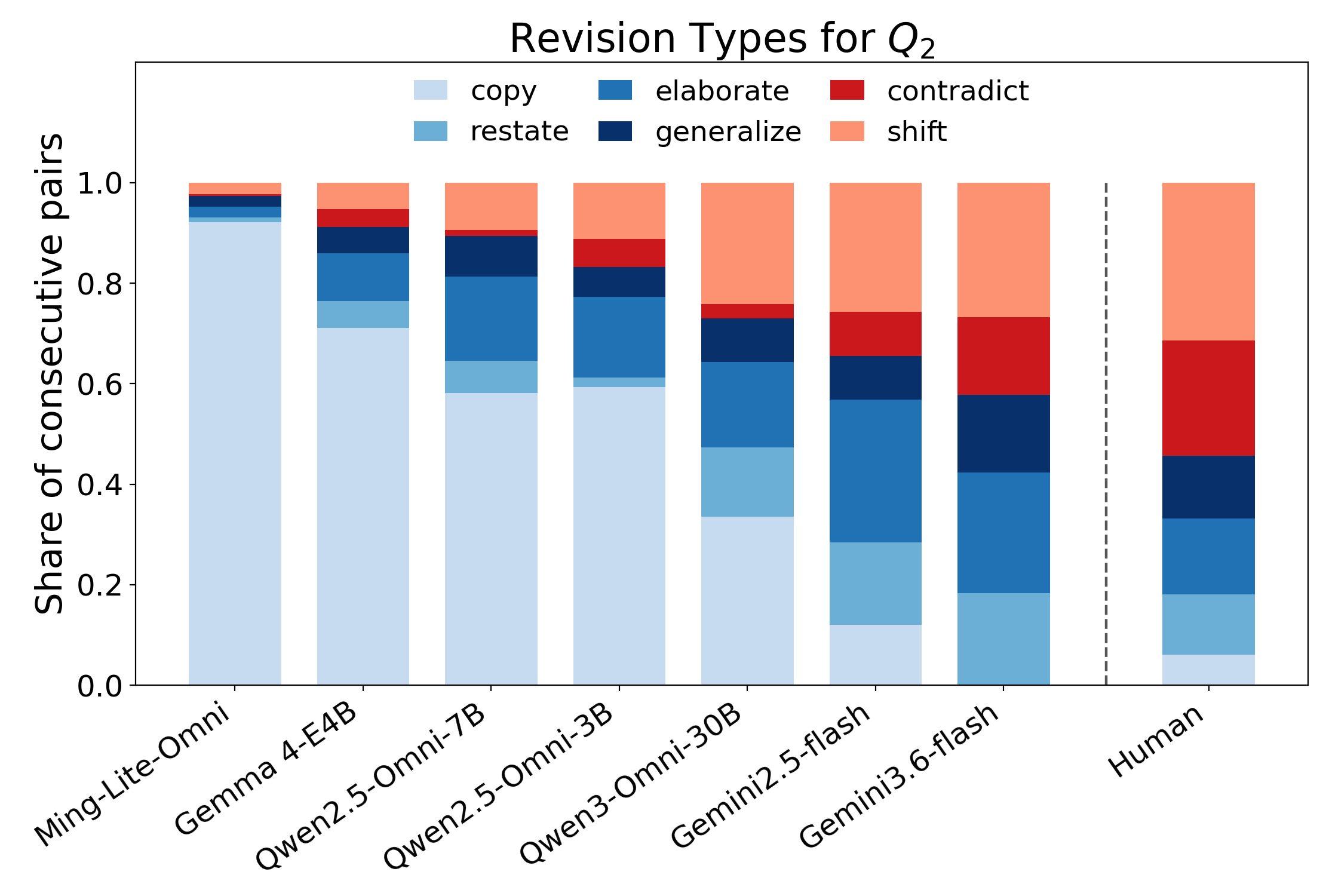}
      \label{fig:revision_q2}
  \end{subfigure}
  \caption{RT classification with NLI-roberta-base model. Humans show notably higher rate of shifting and contradicting than all models, both signs for significant revisions. Many models are occupied with simple copying, and do restating or elaboration more when they do revise.}
  \label{fig:revision}
\end{figure}

\paragraph{Humans revise perceptions more often.}
Classification of RT also separates models and humans as shown in figure~\ref{fig:revision}. Annotators replace their reading (shift/contradict) around $55\%$ of the time on both questions, and copy their previous answer very sparingly. The closest model to human is Gemini3.6-flash at around $45\%$ for both questions, still lower than humans by a gap at $10\%$ level. On the other hand, models other than the two Gemini ones all show a much more repetition manner over adjacent responses: their pairs are often dominated instead by \textbf{copy}, especially for Ming-Lite-Omni which repeats itself at around $85\%$ and $90\%$ for $Q_1$ and $Q_2$, and also includes Qwen2.5-7B/3B and Gemma. In cases where they do move, they \textbf{elaborate} or \textbf{restate} more often: the previous reading is retained and new words are appended or rephrased around it, so the answer moves while the hypothesis is never at risk. In the Appendix~\ref{subsec:revision_details} to~\ref{subsec:rank_gain} we provide more results in support for our conclusion.

We note that frequent revision and early anticipation are not in tension. Anticipation is a rank that identifies which utterance an early reading is about, whereas a contradiction concerns what is claimed about it: a reading can keep its scene and still reverse its claim. Only a reading specific enough to be wrong can be retracted, so frequent revision is the cost of committing early, which humans do but most models don't.


\subsection{Spatial results: gaze manipulation similarity} \label{subsec:spatial_results}

\begin{figure}[ht]
  \centering
  \begin{subfigure}{0.49\textwidth}
      \includegraphics[width=\linewidth]{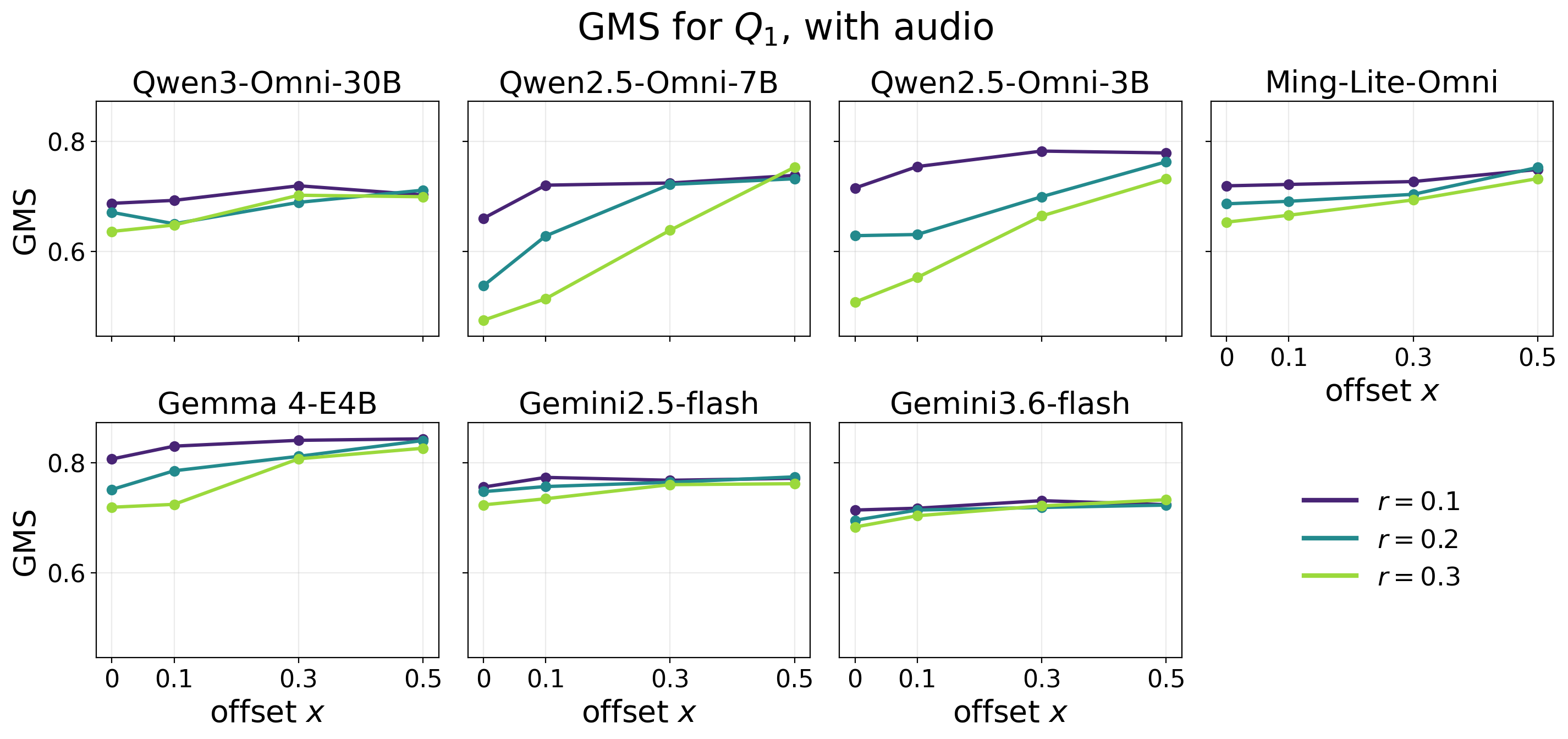}
      \label{fig:gms_q1}
  \end{subfigure}
  \begin{subfigure}{0.49\textwidth}
      \includegraphics[width=\linewidth]{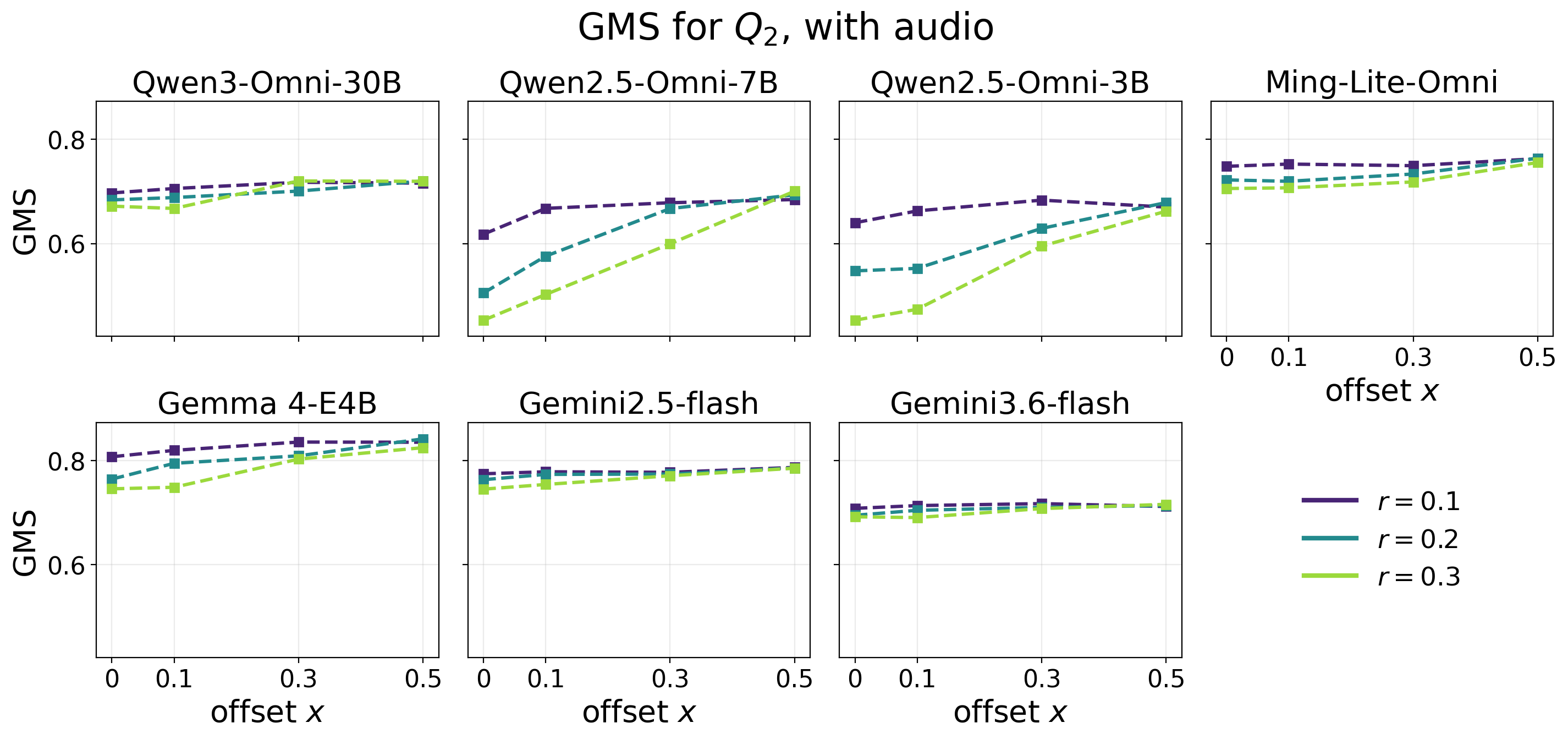}
      \label{fig:gms_q2}
  \end{subfigure}
  \caption{GMS plot for $Q_1$ and $Q_2$. Qwen2.5 models has lower GMS with larger $r$ and smaller $x$, showing they are sensitive to corruption. Gemma shows same pattern for $Q_1$ but not obvious for $Q_2$. Other models are not sensitive to gaze based manipulation. }
  \label{fig:gms}
  \vspace{-1.5em}
\end{figure}

\begin{figure}[ht]
  \centering
  \includegraphics[width=0.8\textwidth]{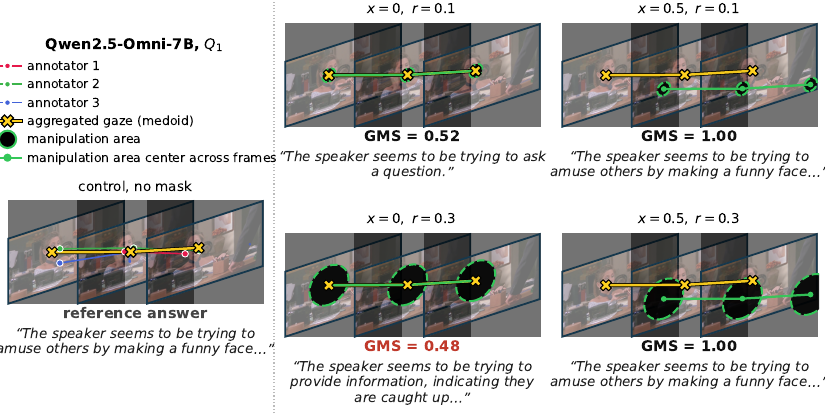}
  \caption{A qualitative illustration of GMS. For Qwen2.5-Omni-7B model on $Q_1$, $(x,r)=(0,0.3)$ gives the lowest GMS, while $(0.5,0.1)$ gives $1$ on GMS, showing its sensitivity over $(x,r)$. Video source from Mintrec2. A complete sample of video manipulation and the answers is provided in the supplementary material.}
  ~\label{fig:gms_qualitative}
  \vspace{-1.5em}
\end{figure}

We compare GMS by $x \in \{0,0.1,0.3,0.5\}$ and $r \in \{0.1,0.2,0.3\}$ against the untouched control. Figure~\ref{fig:gms} shows GMS for $Q_1$ and $Q_2$, and figure~\ref{fig:gms_qualitative} provides a qualitative illustration.

\paragraph{Perception gaps exists where human-attended regions are corrupted.}

We observe that corrupting human-attended regions have an impact on model perception, and it is not only due to corruption itself. In almost all cases, GMS increases as offset $x$ increases. This shows that corrupting human attended regions pushes away the final perception more than corrupting other regions with equally. For some models corrupting more area impacts the perception more. Qwen2.5-Omni-7B is representative: the three lines show a clear gap for different $r$s. However, for some models this effect is not obvious, for example Gemini2.5/3.6-flash, where GMS is nearly flat as $(x,r)$ changes. In~\ref{subsec:gms_no_audio} we also report the results without audio input.


\paragraph{GMS Sensitivity is strongly model-dependent.}
GMS plot pattern over $(x,r)$ is highly influenced by models. The Qwen2.5-Omni models are strongly gaze-sensitive, each spanning roughly $0.2$ of the similarity range on the $Q_2$ grid. Ming-Lite-Omni and the two Gemini models are nearly flat, varying by well under $0.2$ across all results. Note that here flatness on the grid can mean either model robustness or barely reading the visual channel. However, combined with the fact that Ming-Lite-Omni copies its previous answer in $80\%-90\%$ of the time, a flat response is consistent with output that is largely insensitive to input. The Gemini models, on the other hand, copy rarely, have the revision profile closest to humans, so their flat grids are likely showing robustness.

\vspace{-2pt}
\section{Limitations and Conclusion}

\paragraph{Limitations.}
\methodname is limited to short, face-to-face utterances with clearly bounded speakers, drawn from a mixture of scripted and unscripted datasets. Broader settings such as longer conversations, multi-party interactions, or noisier environments, may require additional annotation protocols. Human annotations are collected on a representative subset rather than the full benchmark. Our metrics use sentence-embedding similarity as a scalable proxy for semantic agreement, which may miss subtle pragmatic distinctions between open-ended responses. Finally, human gaze based corruption should be interpreted as a human-grounded corruption-sensitivity test, not as a causal account of human perception or as a saliency estimator.

\paragraph{Conclusion.}
We introduced \methodname, a human-grounded benchmark for evaluating how multimodal LLMs form social interpretations as utterances unfold. Its temporal pipeline uses cumulative sub-utterance clips to measure whether models could make early predictive perceptions in alignment with human understanding, while its spatial pipeline uses human gaze to test model sensitivity to human-attended visual cues. Across several datasets, we find a consistent human-model gap: humans make more specific, and more aligning perceptions with other humans at early stages when they see the video. They also revise more often without being sparing with shifting or contradicting their previous answers. Finally, models also vary significantly on whether their perception is influenced by human gaze patterns. These results suggest that final-answer accuracy is insufficient for evaluating socially interactive agents; models must also form useful provisional interpretations, revise them incrementally, and ground the judgments in communicatively meaningful cues.

\section*{Ethical statement}
In the data collection procedure of \methodname, participants were informed about the study content, and required to sign a consent form, as mentioned in Appendix~\ref{subsec:human_annotation}. There were informed of our measures to store, process, and use their recorded data. No dangerous content or measures appear or is exposed in any of the annotation tasks. 

\section*{Reproducibility statement}
To support reproducibility, we open-source our codes, including data curation procedure, model prompting strategies, and metric analysis, as mentioned in the abstract. 

\typeout{========== ARXIV REACHED BIBLIOGRAPHY ==========}
\bibliography{references}
\bibliographystyle{iclr2027_conference}

\newpage
\appendix

{\Large
\textbf{Appendices} \\}


\section{Data Curation and details} \label{sec:dataset_info}
\subsection{Dataset selection rationale}


We selected source datasets according to four criteria. First, the data should contain face-to-face conversational utterances in which social meaning is expressed through multimodal cues such as speech, facial expression, gaze, etc. Second, utterance boundaries should be sufficiently clear to support cumulative sub-utterance clipping. Third, the data should include short-to-medium utterances, since very long videos make dense human annotation and eye-tracking collection less practical. Fourth, the benchmark should include both controlled and naturalistic interaction settings.

MELD~\citep{poria2019meld} (GPL3.0 license) and MIntRec2~\citep{zhang2024mintrec2} (CC-BY-NC-SA-4.0 license) provide scripted utterances from television dialogue. Although scripted, these clips retain many perceptual demands of everyday social understanding: viewers must integrate lexical, prosodic, facial, gestural, interpersonal, and turn-taking cues to infer speaker intention and possible responses. Prior work also suggests that television dialogue shares important linguistic properties with spontaneous conversation~\citep{quaglio2008television, Bednarek_2018}, while offering cleaner utterance boundaries and greater accessibility for annotation. Also, simulated stimuli has been used for studying human responses in social interactions in neuro-science~\citep{redcay2020social}. This aligns with our goal, as gestalt principles are also highly based on findings from neuro-science side~\citep{koffka2013principles, shamma2011temporal}. We deem that the task of social signal perception is of the similar end for what simulated signals could afford. In fact, these controlled stimuli are frequently used for simulated narratives~\citep{tikka2023narrative, jaaskelainen2021movies} and there are also arguments that they offer good balance by having controlled scenarios as well as decent level of ecological validity~\citep{fan2021levels}. Seamless Interaction~\citep{agrawal2025seamless}(CC-BY-NC-SA-4.0 license) also complements these sources with unscripted dyadic conversations, providing more natural face-to-face interaction. We therefore combine scripted and unscripted sources to balance cue diversity, annotation feasibility, and ecological validity.

We do not include in-the-wild activity-video datasets as primary sources because many clips contain weakly bounded utterances, unclear speaker roles, fewer face-to-face exchanges, or audio/visual conditions that make sub-utterance annotation difficult~\citep{grauman2022ego4d,rossetto2025castle}. Future versions of \methodname could extend the protocol to longer, noisier, multi-party, and more culturally diverse interactions.
\subsection{Annotation setup} \label{subsec:human_annotation} 

\begin{figure}[ht]
  \centering
  \includegraphics[width=\textwidth]{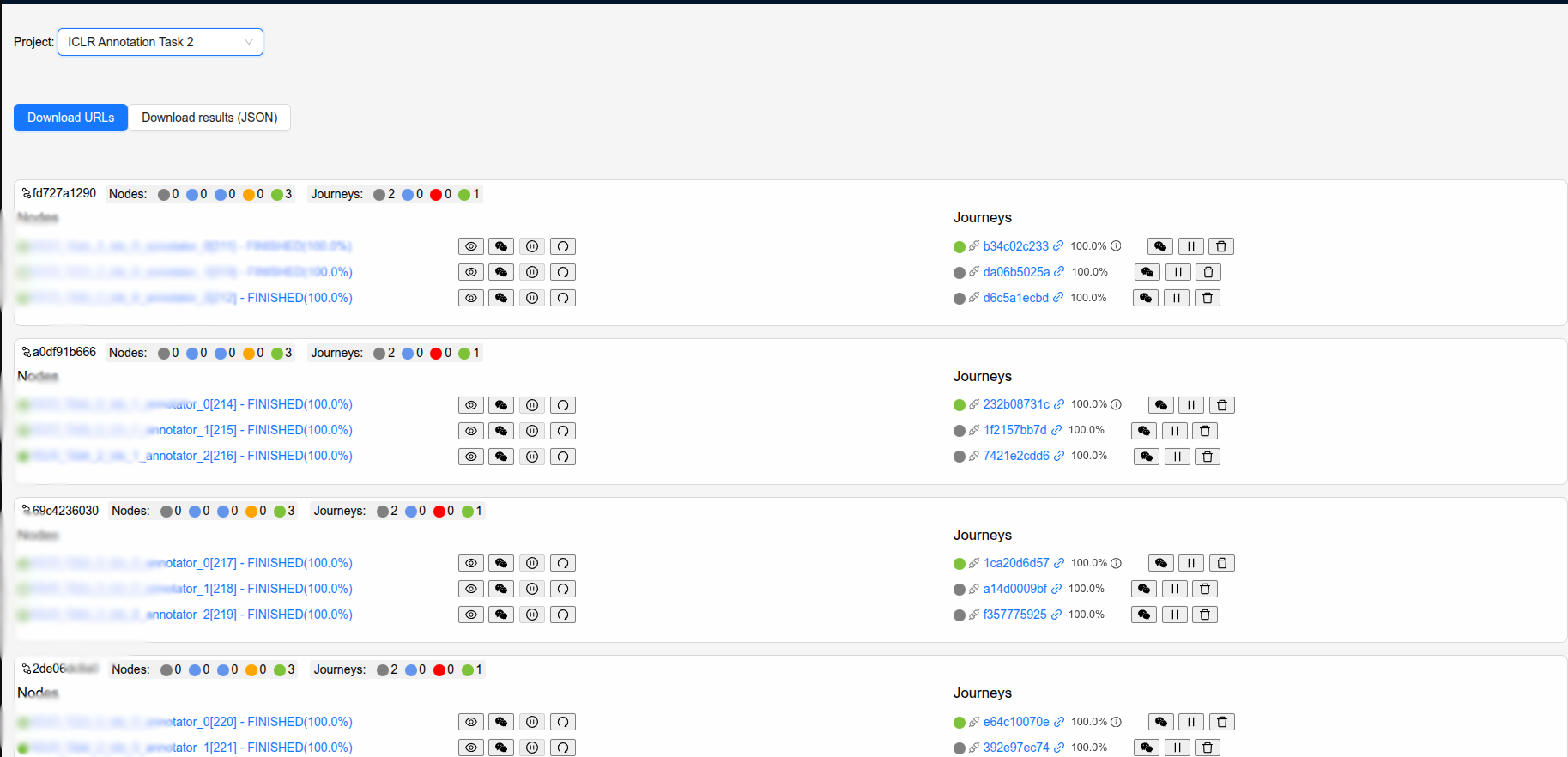}
  \caption{Illustration of the Covfee main page for monitoring the annotation tasks.}
  ~\label{fig:annotation_monitor}
\end{figure}

\methodname records two types of human data: (1) human answers for CVCs: $H_Q(D_1), \ldots, H_Q(D_n)$, and (2) eye-gaze recordings. Both types of responses are recorded by the Covfee tool~\citep{quiros2022covfee}. It provides clear and smooth setup process of annotation tasks. Figure~\ref{fig:annotation_monitor} shows a sample screenshot for monitoring the progress of the annotation tasks.

\begin{figure}[ht]
  \centering
  \begin{subfigure}{0.55\textwidth}
      \includegraphics[width=\linewidth]{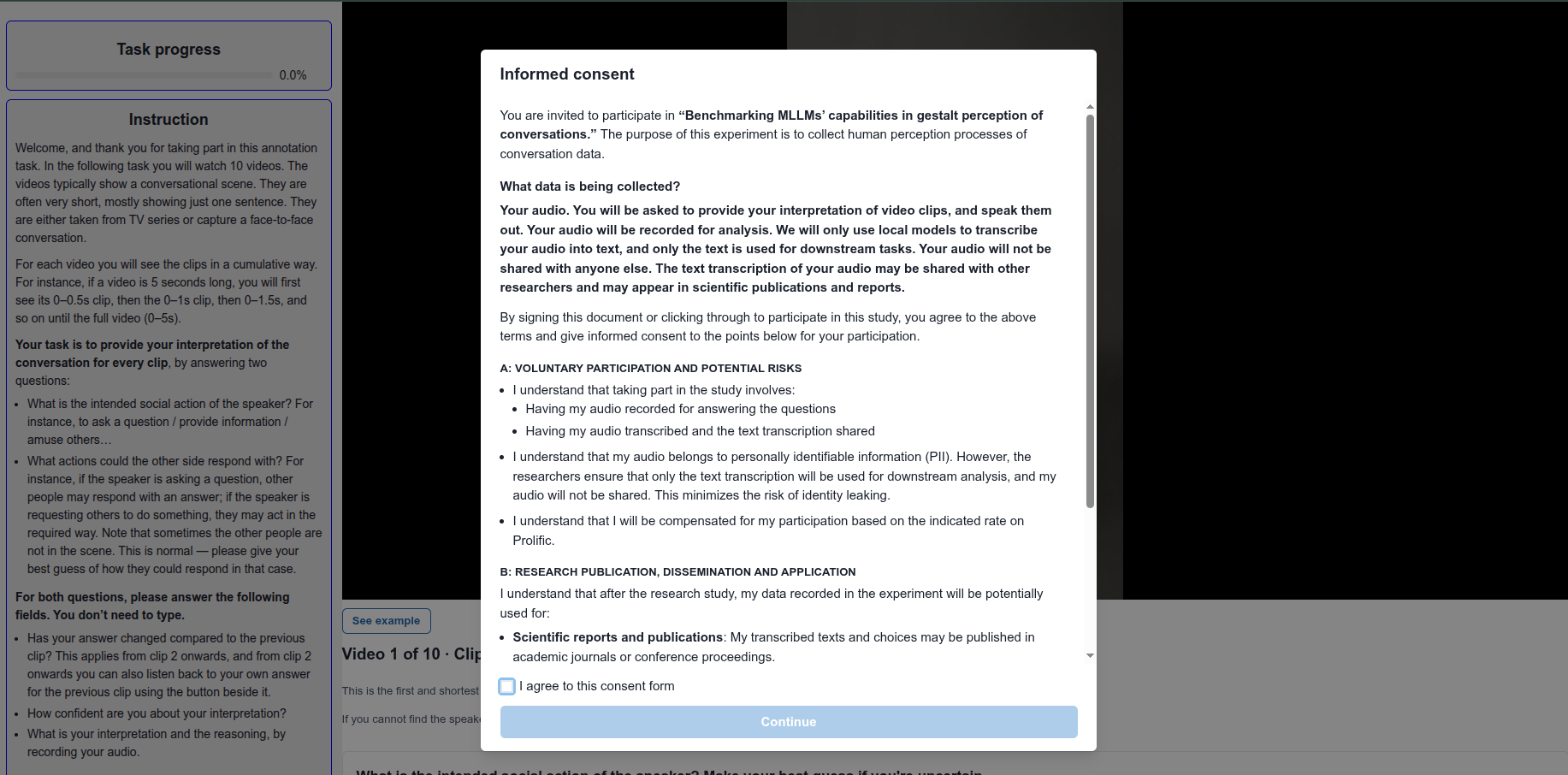}
      \caption{}
      \label{fig:gaze_annotation_consent}
  \end{subfigure}
  \begin{subfigure}{0.55\textwidth}
      \includegraphics[width=\linewidth]{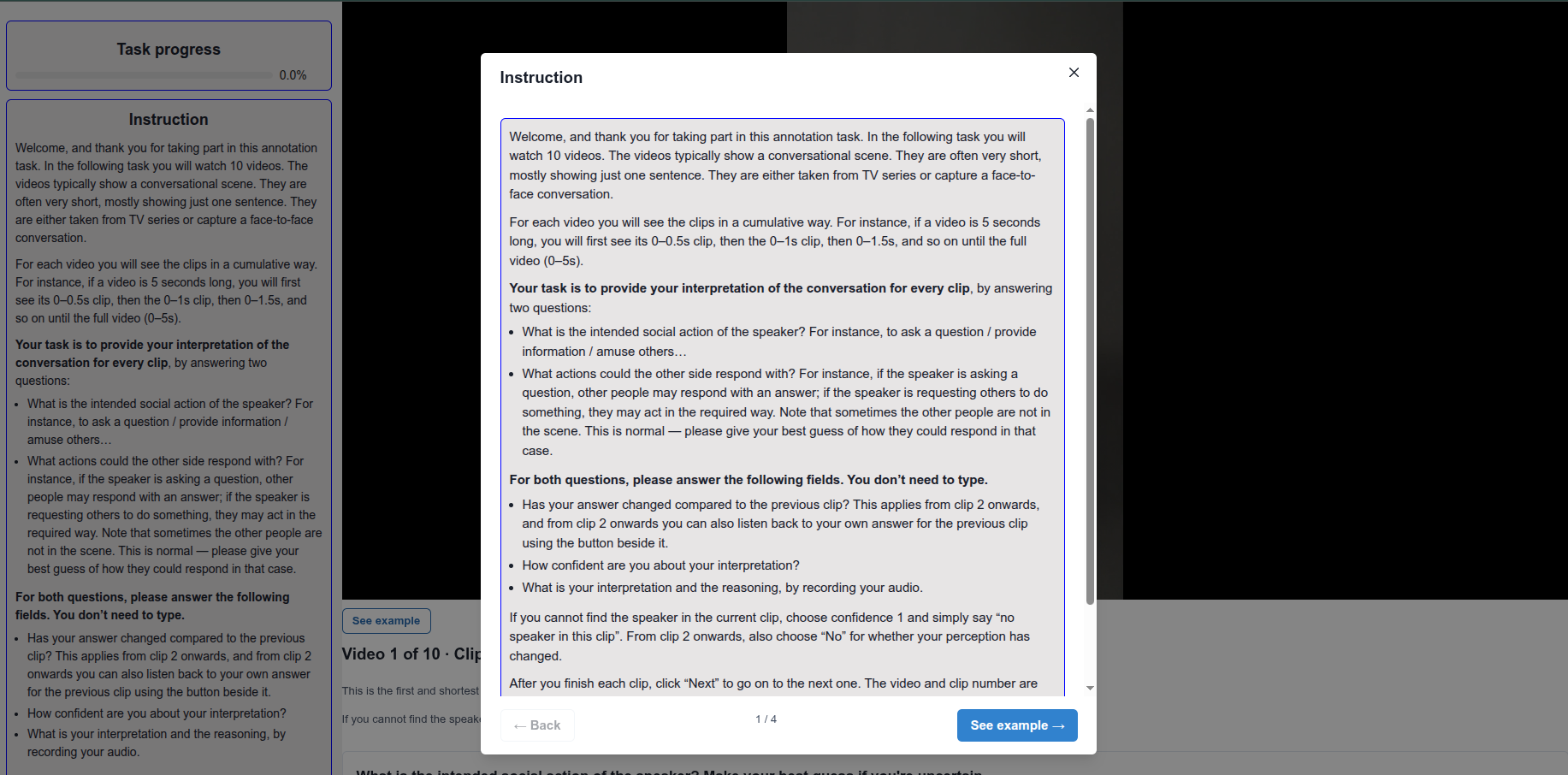}
      \caption{}
      \label{fig:gaze_annotation_instruction}
  \end{subfigure}
  \begin{subfigure}{0.55\textwidth}
      \includegraphics[width=\linewidth]{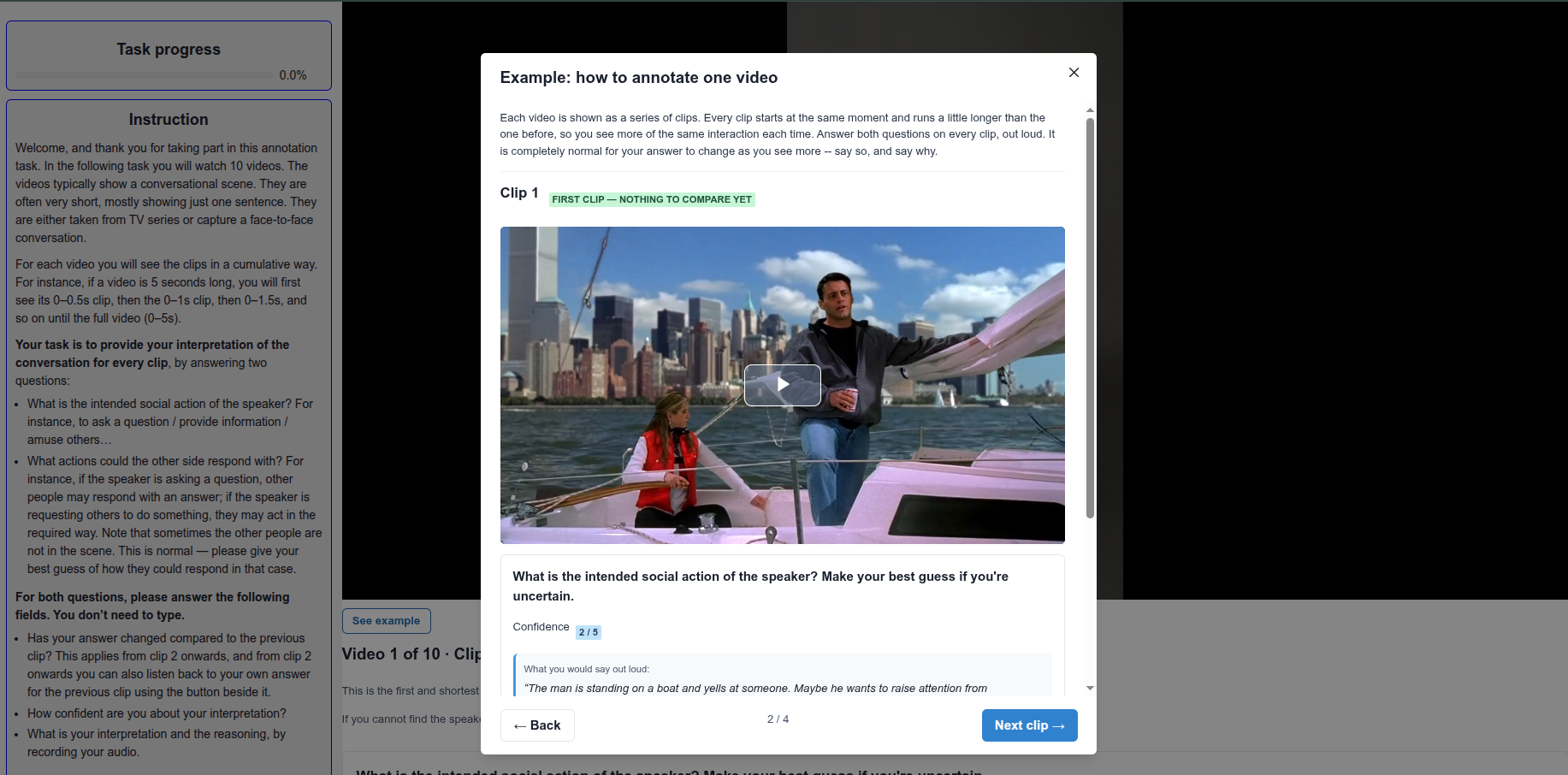}
      \caption{}
      \label{fig:gaze_annotation_example}
  \end{subfigure}
  \begin{subfigure}{0.55\textwidth}
      \includegraphics[width=\linewidth]{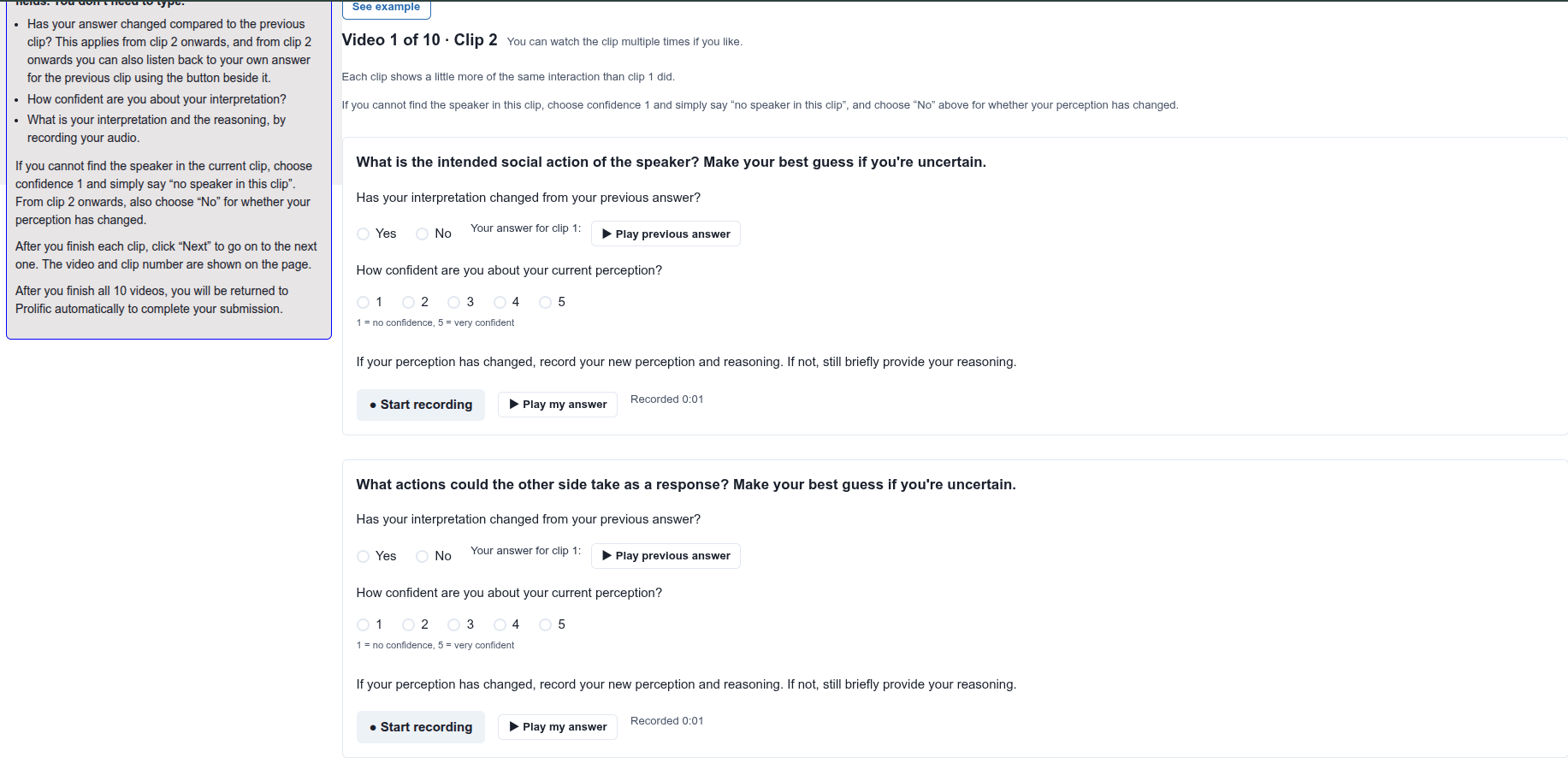}
      \caption{}
      \label{fig:gaze_annotation_panel}
  \end{subfigure}
  \caption{Sample of paid annotators user interface on Covfee when doing the text annotation. They are first asked to sign the consent form to procees(~\ref{fig:gaze_annotation_consent}), then go through the instruction(~\ref{fig:gaze_annotation_instruction}) and example(~\ref{fig:gaze_annotation_example}). ~\ref{fig:gaze_annotation_panel} shows the user actual panel they see during annotation: specifying whether perception has changed, choosing confidence, and recording audios about their perception and reasoning.}
  \label{fig:text_annotation_ui}
\end{figure}

\subsubsection{Human text annotations}
We employ a crowd sourcing approach for collecting human text annotations on CVC. We utilize the Prolific~\footnote{https://app.prolific.com/}, and provided them with a link to the Covfee server which is set up locally. Every annotator sign a consent form before taking part in. During the task, we asked the annotators to provide the following for both $Q_1$ and $Q_2$:
\begin{itemize}
    \item Their perception and reasons over the current clip $D_k$, by speaking it out and audio recorded;
    \item Their current confidence, with a scale from 1-5;
    \item Whether their perception has changed compared to the previous clip ($D_{k-1}$, only applies to clips with $k\geq 2$)
\end{itemize}
and a screenshot of the Covfee page for the annotators is shown in figure~\ref{fig:text_annotation_ui}. After collection, their audio are transcribed into text using WhisperX~\citep{bain2022whisperx} model. In practice the implementation is done with fast WhisterX~\footnote{https://github.com/SYSTRAN/faster-whisper}.

\subsubsection{Human eye gaze data}
To collect annotators' eye focus when they watch a video, we utilize the Pupil Core Eye Tracking Glasses~\footnote{https://pupil-labs.com/products/core}. It takes videos for people's eyeballs and then infers their eye gazes in the world camera. If a surface is provided and detected in the scene, it can automatically calibrate and calculate relative positions of eye gaze on the specific surface. Since we also have fixed region of the video display area on the screen, we can then infer the eye gaze position within the video region. Annotators sign a consent form informing them about the fact that their eye gaze data will be processed and published before they annotate.

\begin{figure}[ht]
  \centering
  \includegraphics[width=\textwidth]{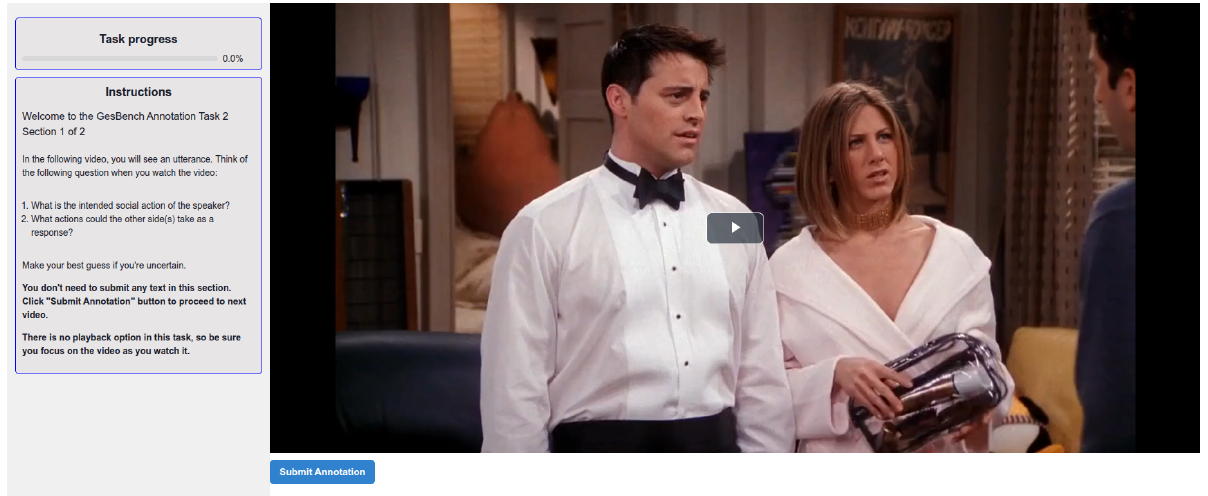}
  \caption{Example of the annotator task page for providing gaze annotations. The left is an instruction panel, and the right shows the video. The annotators are asked to think of the two questions ($Q_1$, $Q_2$) during watching, and they click "Submit Annotation" button to proceed to the next video. The scrolling bar is intentionally disabled for this task to ensure that the video displaying region stays fixed within the screen.}
  ~\label{fig:gaze_annotation}
\end{figure}

We use AprilTags provided on the official site of Pupil~\footnote{https://docs.pupil-labs.com/core/software/pupil-capture/\#surface-tracking} to mark the computer screen in the scene of the world camera in Pupil Core. All annotators did a calibration session~\footnote{https://docs.pupil-labs.com/core/software/pupil-capture/\#calibration-process} before they started the experiment. During the experiment, they were asked to think of two questions, $Q_1$ and $Q_2$, during watching the videos, and then they move on to the next video. There is no video cutting in this experiment, as only continuous gazing during watching an entire video would enable us to construct a per-frame gaze-position on the videos. An example layout of one annotation page is provided in figure~\ref{fig:gaze_annotation}.

\subsection{Dataset details} \label{subsec:dataset_info}

\begin{table}[ht]
\centering
\begin{tabular}{lrrrrrrrrrr}
\toprule
 & \multicolumn{2}{c}{\textbf{Full}} & \multicolumn{2}{c}{$\mathcal{T}_1$} & \multicolumn{2}{c}{$\mathcal{T}_2$} & \multicolumn{2}{c}{$\mathcal{T}_3$} & \multicolumn{2}{c}{$\mathcal{S}$} \\
\cmidrule(lr){2-3} \cmidrule(lr){4-5} \cmidrule(lr){6-7} \cmidrule(lr){8-9} \cmidrule(lr){10-11}
Dataset & \# utt. & AL (s) & \# utt. & AL & \# utt. & AL & \# utt. & AL & \# utt. & AL \\
\midrule
MELD & 1,870 & 3.273 & 216 & 3.229 & 141 & 3.284 & 8 & 2.938 & 219 & 3.242 \\
MIntRec2 & 2,974 & 2.968 & 215 & 2.928 & 143 & 2.871 & 8 & 2.500 & 219 & 2.922 \\
Seamless & 1,364 & 4.117 & 159 & 4.157 & 105 & 4.171 & 4 & 5.000 & 162 & 4.154 \\
\midrule
Overall & 6,208 & 3.312 & 590 & 3.369 & 389 & 3.371 & 20 & 3.175 & 600 & 3.372 \\
\bottomrule
\end{tabular}
\caption{\methodname data details. Seamless means Seamless Interaction. AL means average utterance length.}
\label{tab:data_stats}
\end{table}

We show the full details of our dataset in table~\ref{tab:data_stats}. $\mathcal{T}_1,\mathcal{T}_2,\mathcal{T}_3$ represent the subset with \textit{at least} $1/2/3$ text annotators respectively, so they are subsuming instead of exclusive. $\mathcal{S}$ denotes the set with gaze annotations, all clips have $3$ annotators.

In the experiments, $\mathrm{Spec}$ and $\mathrm{Ant}$ are measured on $\mathcal{T}_2$, since the $\mathrm{Ant}$ needs at least two annotators to compute human value; RT is done on $\mathcal{T}_1$. GMS is measured on $\mathcal{S}$. Due to API costs, both Gemini models are prompted on only $\mathcal{T}_1$ and $\mathcal{S}$.

\section{Experimental Setup} \label{sec:experimental_setup}

\subsection{Model prompting} \label{subsec:model_prompting}
In this subsection we provide details for configuring and prompting the models. For all models, we employ a zero-shot strategy, and each video clip in CVC and corrupted video sets are prompted separately by launching a new conversation.  all models, the user inputs within the chat template are as follows:

\begin{promptbox}{GESBENCH PROMPT (full, for CVC clips $\{D_k\}, k=1$)}
Watch the following video. You will see it in a cumulative way: all clips are cut from the same single source video. For instance, if a video is 5 seconds long, you will first see its 0–0.5s clip, then the 0–1s clip, then 0–1.5s, and so on until the full video (0–5s).

The current and first video clip is given in the media. Answer the following two questions:

Q1: What is the intended social action of the speaker? For instance, to ask a question / provide information / amuse others, etc.
- State your interpretation.
- How confident are you about your interpretation? Answer with a number in the scale of 1-5. 1 is no confidence and 5 is very confident.

Q2: What actions could the other side respond with? For instance, if the speaker is asking a question, other people may respond with an answer; if the speaker is requesting others to do something, they may act in the required way.
- State your interpretation.
- How confident are you about your interpretation? Answer with a number in the scale of 1-5. 1 is no confidence and 5 is very confident.
\end{promptbox}

\begin{promptbox}{GESBENCH PROMPT (full, for CVC clips $\{D_k\}, k\geq 2$)}
This is the <clip_ordinal> clip, cut from the same source video as the previous clips. Your answer for previous clip was: <previous_answer> 
The current video clip is given in the media. Regarding current clip, still think of the two questions:

Q1: What is the intended social action of the speaker? For instance, to ask a question / provide information / amuse others, etc.
Q2: What actions could the other side respond with? For instance, if the speaker is asking a question, other people may respond with an answer; if the speaker is requesting others to do something, they may act in the required way.

For each of Q1 and Q2, answer the following questions separately:
- Has your perception changed compared to previous clip? Say explicitly "yes" or "no".
- If yes, state your new perception, and provide reasons. If no, also provide your reasons.
- How confident are you about your interpretation now? Answer with a number in the scale of 1-5. 1 is no confidence and 5 is very confident.
\end{promptbox}


where \texttt{<previous\_answer>} tag is for the model's answer at the previous round. The prompt is done in a multi-turn way, ensuring that models have the full context of their answer previously and in the same way as humans, since human would naturally remember what they have just seen. 

Below we list some necessary parameters used when prompting:
\paragraph{Qwen2.5-Omni-7B/3B.} We follow the official chat template provided on HuggingFace~\footnote{https://huggingface.co/Qwen/Qwen2.5-Omni-7B}~\footnote{https://huggingface.co/Qwen/Qwen2.5-Omni-3B}, where it includes fields for video and audio inputs. The model weights and configurations are downloaded from the above HuggingFace repositories respectively. All of the model inference is done on Intel Xeon Gold 6448Y 32C 2.1GHz CPU and NVIDIA A100 GPU(80G memory), with CUDA version 12.6. 

\paragraph{Qwen3Omni30B.} We follow the official chat template provided on HuggingFace~\footnote{https://huggingface.co/Qwen/Qwen3-Omni-30B-A3B-Instruct}. The model weights and configurations are downloaded from the above HuggingFace repositories. All of the model inference is done on AMD EPYC 9334 2.7 GHz
210W and NVIDIA H100 GPU(80G memory), with CUDA version 12.8 and PyTorch version 2.7.1. 

\paragraph{Ming-Lite-Omni.}We follow the official chat template provided on HuggingFace~\footnote{https://huggingface.co/inclusionAI/Ming-Lite-Omni}. The model weights and configurations are downloaded from HuggingFace repository as above. The inference for CVC is done on Intel Xeon Gold 6448Y 32C 2.1GHz CPU and NVIDIA A100 GPU, with CUDA version 12.8, and the inference for GMS is done on Intel Xeon Gold 5128 CPU and NVIDIA RTX 6000 Pro GPU(96G memory), with CUDA version 12.8.

\paragraph{Gemma 4.} We follow the chat guidlines provided by Google~\footnote{https://ai.google.dev/gemma/docs/core}. We download the models weights and configurations from HuggingFace~\footnote{https://huggingface.co/google/gemma-4-E4B}. Max-frame is set to 32 and thinking is disabled. All of the model inference is done on AMD EPYC 7413 CPU and NVIDIA A40 GPU(48G memory), with CUDA version 12.6.

\paragraph{Gemini2.5/3.6-Flash.} We implement according the official documentation provided~\footnote{https://ai.google.dev/gemini-api/docs/models/gemini-2.5-flash}. The evaluations are done through API calls where we request batch processing.



\subsection{Answer Postprocessing} \label{subsec:anwer_processing}
We postprocess the answers from both the models and humans. 

\paragraph{Model answer extraction.} For models, we employ a multi-turn conversation manner: prompting them with a media and their previous answers. This is intended to match the human experiment environment, where humans can naturally remember what they have actually seen. Since humans are answering both $Q_1$ and $Q_2$ under the same context, their answers maybe naturally associated. To ensure maximum comparability, we also prompt the models with the two question combined (see~\ref{subsec:model_prompting}) instead of separately, and then use another LLM to extract both fields. We asked the model to extract both fields verbatim as much as possible and not adding, deleting, summarizing, or modifying the content. Specifically, we leverage OpenAI's GPT-4o-Mini~\footnote{https://developers.openai.com/api/docs/models/gpt-4o-mini} by API call, with the following prompt for clips of $k=1$ and $k\geq 2$ respectively:

\begin{promptbox}{GESBENCH ESTRACTION PROMPT for CVC clips $\{D_k\}, k=1$}
You are extracting structured fields from an answer another model gave about the FIRST clip of a video. That model was shown the video cumulatively -- clip 1 is the opening fragment -- and was asked two questions about it:

Q1 (intention): What is the intended social action of the speaker?
Q2 (response): What actions could the other side respond with?

The user message is that model's answer, verbatim. Report what it said about each question, as JSON.

- "changed" is always null here. This is the first clip, so there is nothing for a perception to have changed from.
- "confidence" is the 1-5 number the model gave for that question, as a number. Use null if it gave none for that question.
- "perception" is everything the model wrote in answer to that question, copied verbatim: the same words in the same order, including its perception, reasons, hedges and alternative readings. Do not paraphrase, summarise, shorten, correct or reorder it, and do not add anything of your own. Leave out only the question label or a restatement of the question ("Q1:", "Intended social action:") and the confidence rating itself. If one passage answers both questions together and cannot be split, copy it into both. Use "" if it said nothing about that question.
- "insufficient_information" is true only when the model declines to answer that question outright -- "there is not enough information", "I don't know", "it is impossible to say from this clip" -- and offers no interpretation alongside saying so. Record that statement as the "perception" when it happens, so that declining to answer is never confused with an answer nobody extracted. A tentative, hedged or low-confidence guess is still a guess: false, however unsure the model sounds. An answer that simply never mentions the question is false as well -- saying nothing is not the same as saying it cannot be known.
- Report only what is in the answer. If the answer is empty, is an error message, or does not address one of the questions, use null and "" for that question rather than guessing.
- The two questions are usually labelled Q1 and Q2, but not always. Match on what is being answered -- the speaker's own intent goes in "intention", what someone else might do in reply goes in "response" -- rather than on the labels.
\end{promptbox}

\begin{promptbox}{GESBENCH ESTRACTION PROMPT for CVC clips $\{D_k\}, k\geq 2$}
You are extracting structured fields from an answer another model gave about a later clip of a video. That model was shown the video cumulatively, had already answered about the previous clip, and was asked, for each of two questions, whether its perception had changed, what its perception is now, and how confident it is:

Q1 (intention): What is the intended social action of the speaker?
Q2 (response): What actions could the other side respond with?

The user message is that model's answer, verbatim. Report what it said about each question, as JSON.

- "changed" is the model's own explicit "yes" or "no" for that question, lowercased. If it does not say so in those words, take it from whether it describes its perception as revised or as unchanged. Use null only when the answer does not address that question at all.
- "confidence" is the 1-5 number the model gave for that question, as a number. Use null if it gave none for that question.
- "perception" is everything the model wrote about that question for this clip, copied verbatim: the same words in the same order, including its perception, reasons, hedges and alternative readings. Do not paraphrase, summarise, shorten, correct or reorder it, and do not add anything of your own. When it says its perception has not changed, copy what it wrote anyway -- the restated interpretation and the reasons alike. Leave out only the question label or a restatement of the question ("Q1:", "Intended social action:"), the confidence rating itself, and a bare "yes" or "no" that does nothing but answer whether the perception changed. If one passage answers both questions together and cannot be split, copy it into both. Use "" only if it wrote nothing about that question.
- "insufficient_information" is true only when the model declines to answer that question outright -- "there is still not enough information", "I don't know", "it is impossible to say from this clip" -- and offers no interpretation alongside saying so. Record that statement as the "perception" when it happens, so that declining to answer is never confused with an answer nobody extracted. A tentative, hedged or low-confidence guess is still a guess: false, however unsure the model sounds. An answer that simply never mentions the question is false as well -- saying nothing is not the same as saying it cannot be known.
- Report only what is in the answer. If the answer is empty, is an error message, or does not address one of the questions, use null and "" for that question rather than guessing.
- The two questions are usually labelled Q1 and Q2, but not always. Match on what is being answered -- the speaker's own intent goes in "intention", what someone else might do in reply goes in "response" -- rather than on the labels.
\end{promptbox}

In the GMS tests, since we are comparing perceptions, we ask the extraction model to drop reasoning and just kept the perception itself. The difference between extraction prompt does not invalidate GMS test as a model's answers are always compared to its own, and no human answers is involved in this process. Prompt is as below:

\begin{promptbox}{GESBENCH ESTRACTION PROMPT for GMS test $\widetilde{D}_n(x,r)$}
You are extracting structured fields from an answer another model gave about a video. That model was shown the whole clip once and was asked two questions about it:

Q1 (intention): What is the intended social action of the speaker?
Q2 (response): What actions could the other side respond with?

The user message is that model's answer, verbatim. Report what it said about each question, as JSON.

- "confidence" is the 1-5 number the model gave for that question, as a number. Use null if it gave none for that question.
- "perception" is the model's own interpretation for that question, in its words or closely paraphrased. Do not add your own reading of the video, do not compress it to a one-word label, and leave out its supporting reasons unless the interpretation cannot be stated without them. Use "" if it said nothing about that question.
- "insufficient_information" is true only when the model declines to answer that question outright -- "there is not enough information", "I don't know", "it is impossible to say from this video" -- and offers no interpretation alongside saying so. Record that statement as the "perception" when it happens, so that declining to answer is never confused with an answer nobody extracted. A tentative, hedged or low-confidence guess is still a guess: false, however unsure the model sounds. An answer that simply never mentions the question is false as well -- saying nothing is not the same as saying it cannot be known.
- Report only what is in the answer. If the answer is empty, is an error message, or does not address one of the questions, use null and "" for that question rather than guessing.
- The two questions are usually labelled Q1 and Q2, but not always. Match on what is being answered -- the speaker's own intent goes in "intention", what someone else might do in reply goes in "response" -- rather than on the labels.
\end{promptbox}

\paragraph{Human answer processing}
Human answers are naturally separated in the two fields so they don't experience this extraction process. With the transcribed human text answer for both questions, we employ a \textit{carry-forward} operation. Sometimes humans say ``no, nothing has changed'' or similar words when their perception did not change, and in this case it can be misleading as the pure linguistic meaning is different from the true perception. We detect this pattern and carry their previous answers forward to replace the explicit indication that their perception did not change in such cases.

We argue, however, that this operation does not significantly impact our methodologies. For $\mathrm{Spec}$ and $\mathrm{Ant}$, the answers are compared to either the same perceiver's own ones at different stages, or another human annotator's final answer. This also makes RT a conservative comparison instead of an inflated one: carry-forward is actually making \textit{copy} patterns for humans, and humans are still having larger proportion of shift/contradict and very few copies even after it.

\subsection{Gaze manipulation details}
Every video in $\mathcal{S}$ has gaze from 3 annotators, recorded with Pupil Core. Samples with confidence below $0.6$ or that fall outside the video area are dropped. For each frame of a clip, each annotator's gaze position is the per-axis median of their samples within $\pm 0.1$\,s of the frame time; if there are none, their nearest sample within $0.5$\,s is used. The frame's gaze point is the \textit{medoid} of the annotators' positions, i.e. the one with the smallest summed distance to the others, so it is always a point an annotator actually looked at. Frames on which no annotator has a position are not masked.

The disc has radius $r\cdot\min(w,h)$. For $x>0$, the offset direction is fixed per clip: we count which quadrant the gaze point falls in over all frames and move the disc towards the opposite corner (gaze mostly top-right is shifted to the bottom-left, and so on), by $xw$ horizontally and $xh$ vertically. If the shifted center stays inside the frame, a disc that crosses the frame edge is cut off there. If it would leave the frame along one axis, the shift along that axis is reversed, placing the disc on the other side of the gaze point.

\section{Additional results} \label{sec:additional_results}


\subsection{Extraction fidelity} \label{subsec:extraction_fidelity}
\begin{table}[ht]
\centering
\begin{tabular}{lrrrrrr}
\toprule
Model & Answers & Coverage & Exact & Cov.\,$<$0.9 \\
\midrule
Qwen3-Omni-30B & 72735 & 0.999& 0.85 & 0.00 \\
Qwen2.5-Omni-7B & 44656 & 1.000& 0.99 & 0.00 \\
Qwen2.5-Omni-3B & 40659 & 0.955& 0.89 & 0.05 \\
Ming-Lite-Omni & 80142 & 0.999 & 0.99 & 0.00 \\
Gemma 4-E4B & 80230 & 1.000 & 1.00 & 0.00 & \\
Gemini2.5-flash & 11506 & 1.000 & 0.96 & 0.00  \\
Gemini3.6-flash & 11406 & 1.000 & 0.96 & 0.00 \\
\bottomrule
\end{tabular}
\caption{Extraction fidelity, with $Q_1$ and $Q_2$ answers pooled. Coverage: mean share of an extracted answer's words found, in order, in the model's reply. Exact: share of answers that are one unbroken stretch of the reply. Cov.\,$<$0.9: share of answers with coverage below 0.9. }

\label{tab:extraction_fidelity}
\end{table}

To verify that our extraction strategy keeps loyalty to the original content, we do a fidelity check on the raw answers and extracted ones, and show the statistics in table~\ref{tab:extraction_fidelity}. We conclude that the extractor copies rather than rewrites. Mean coverage is at least 0.999 for six of the seven models, and at least 85\% of answers are exact.

\subsection{Early commitment profile} \label{subsec:early_profile}

Human comprehension is incremental~\citep{schlangen2011general} and predictive~\citep{pickering2018predicting}: listeners commit to an interpretation before the evidence licenses it~\citep{holler2019multimodal}. 

To further understand the nature in early responses for models and human, we tag every answer with five non-exclusive features by regular-expression matching on the lowercased text, whole words only, using the keywords below:

\textbf{undecided}: \textit{``can't / cannot / unable to / not able to / impossible to / not possible to / hard to / difficult to / too early to''} followed, up to two words later, by \textit{``tell / say / determine / know / judge / interpret / infer / see''}; or \textit{``not sure, unsure, no idea, don't know, do not know, unclear, not clear, insufficient, not enough, nothing yet, nothing to go on''};

\textbf{hedged}: Referring to hedge detection~\citep{lakoff1973hedges,farkas2010conll}, we choose \textit{``might, may, could, possibly, perhaps, maybe, probably, likely, seem(s), appear(s), suggest(s), potentially, guess''};

\textbf{alternatives}: \textit{``or, either, such as, whether''};

\textbf{speech}: \textit{``saying / says / said / tells / telling / told / replies / replying''}, \textit{``is / was / are / were / 's / 're like,''}, or a quotation mark. It only counts in a clause (split on \textit{``. ! ? ;''} or a line break) that contains no prediction marker, so ``he will say `get out' '' counts as prediction, not speech;

\textbf{prediction}: \textit{``will / 'll / won't / shall / gonna''}; \textit{``going to / about to / set to''} followed by a word other than an article or pronoun (so not ``going to the door''); \textit{``expect(s) / expecting / anticipate(s) / anticipating / predict(s) / predicting''}; \textit{``soon / shortly / afterward(s)''}; and \textit{``next''} followed by \textit{``he / she / they / the''}.

We define $f\big(A_Q(D_k)\big)$ to represent whether $A_Q(D_k)$ belongs to category $f$, with $A=H$ or $M$ to represent either human or model answer. Let $\mathcal{K}_\delta(V) = \{k : k/n(V) \le \delta\}$ be the \emph{early window} of an utterance. The early rate is
\begin{equation}
  \mathrm{Rate}_f(A,V) = \frac{1}{|\mathcal{K}_\delta(V)|}\sum_{k \in \mathcal{K}_\delta(V)} \mathds{1}\!\left[f\Big(A_Q\big(D_k(V)\big)\Big)\right].
\end{equation}
We compute it on two texts: the \emph{extracted} answers used everywhere else, and the \emph{unextracted} text, which is the model's stored reply as generated and the annotator's two answers joined.

\begin{figure}[ht]
  \centering
  \includegraphics[width=0.8\textwidth]{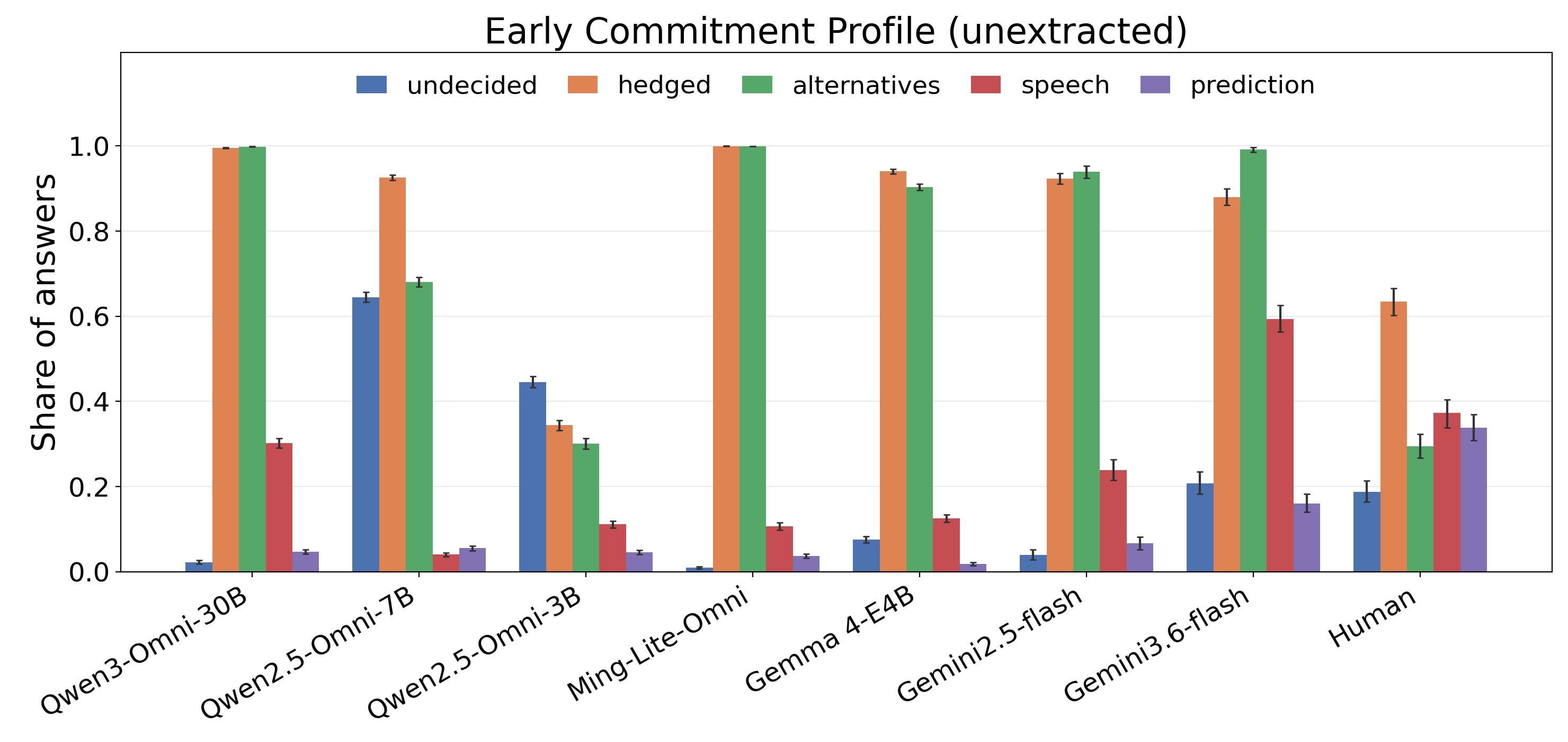}
  \caption{Comparison of early commintment type over categories, with raw texts from models without extraction. Humans are the highest in mentioning predictions, with most models providing hedges or saying there is insufficient information.}
  ~\label{fig:early_profile_blob}
\end{figure}

\begin{figure}[ht]
  \centering
  \begin{subfigure}{0.49\textwidth}
      \includegraphics[width=\linewidth]{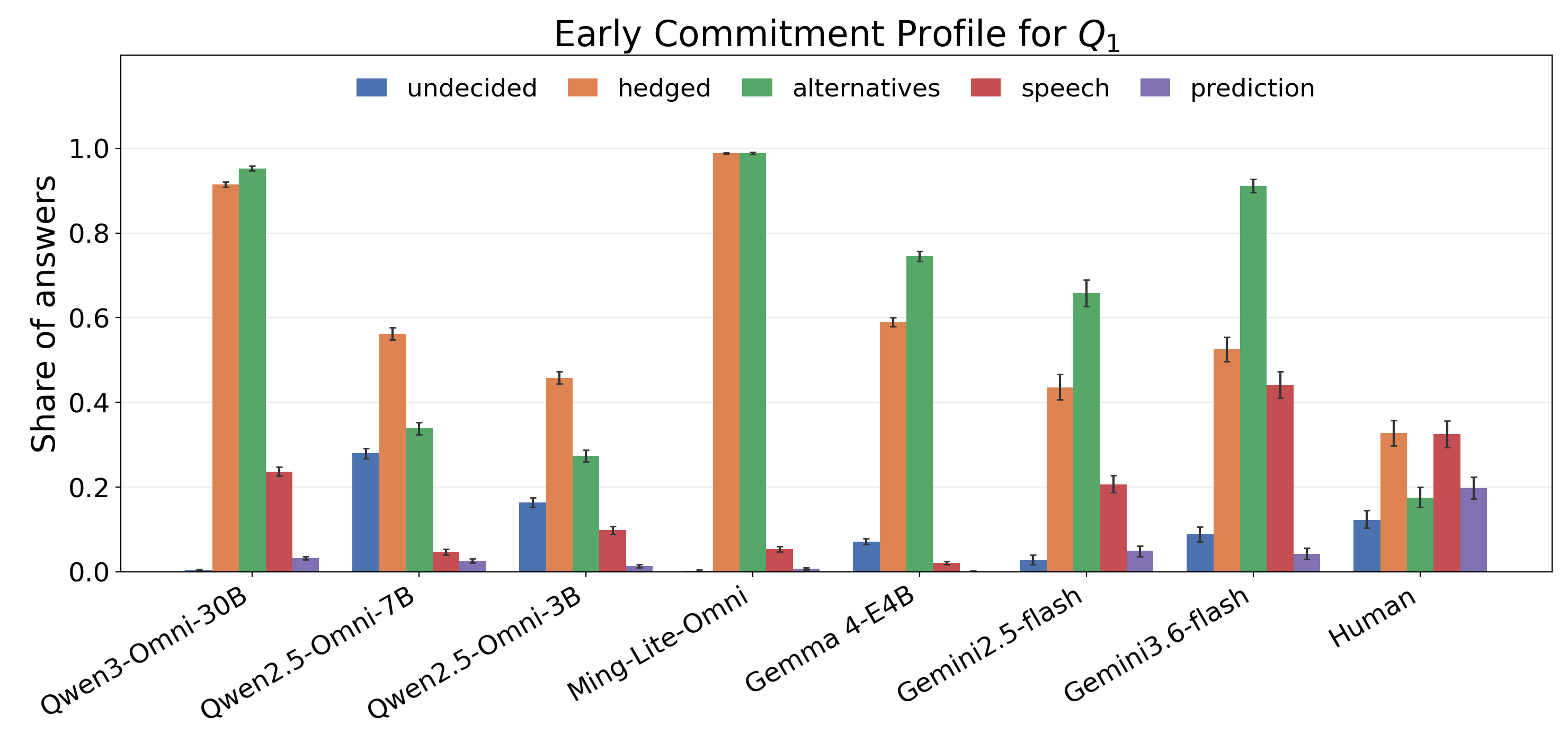}
      \label{fig:early_profile_extracted_q1}
  \end{subfigure}
  \begin{subfigure}{0.49\textwidth}
      \includegraphics[width=\linewidth]{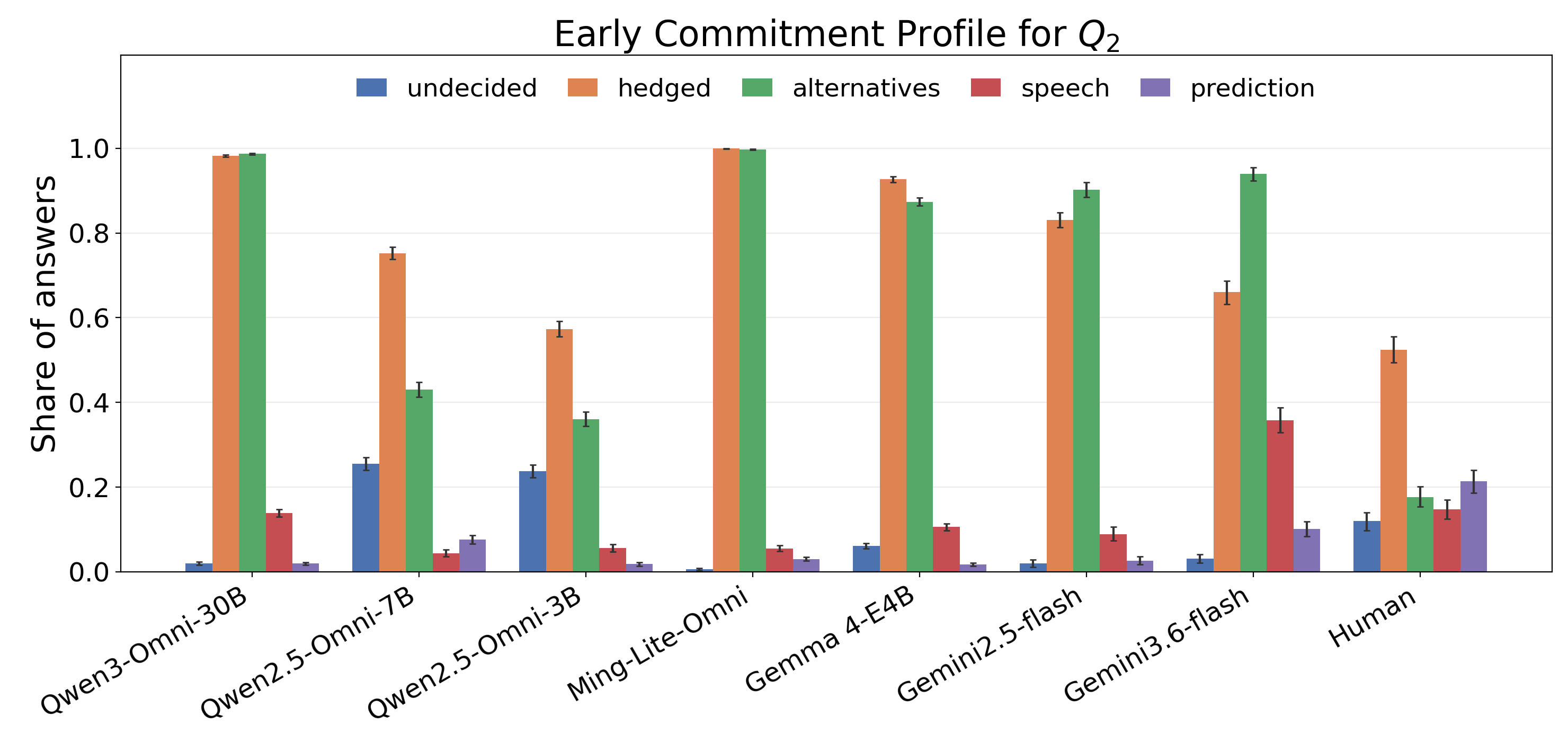}
      \label{fig:early_profile_extracted_q2}
  \end{subfigure}
  \caption{Comparison of early commintment type over categories, on extracted texts for models with both questions. Humans are the highest in mentioning predictions, with most models providing hedges or saying there is insufficient information.}
  \label{fig:early_profile_extracted}
\end{figure}

Figure~\ref{fig:early_profile_blob} shows the unextracted profile, computed on \textbf{full set} when applicable (local models), and on $\mathcal{T}_1$ for Gemini models and humans. Annotators predict what comes next in around $35\%$ of early answers; no model exceeds $18\%$ (Gemini3.6-flash being max), and the other six stay below $10\%$. Models instead hold readings open: Ming-Lite-Omni and Qwen3-Omni-30B offer alternatives and hedge in almost every early answer, and Gemini2.5/3.6-flash and Gemma offer alternatives in $90\%$ to $99\%$, against around $30\%$ for annotators. The Qwen2.5-Omni models fail differently: they are undecided in around $50\%$ (3B) and $60\%$ (7B) of early answers against $20\%$ for annotators. The per-question profiles on the extracted text (Figure~\ref{fig:early_profile_extracted}) show the same ordering: above $20\%$ human predictions on $Q_1$ against at most $6\%$ for any model. Both behaviors run against the Gestalt account: a perceiver who lists the possibilities has not closed on a reading, and one who declines has not begun.

\subsection{Revision classification details} \label{subsec:revision_details}

We use the pre-trained cross-encoder \texttt{nli-roberta-base}~\citep{liu2019roberta} without fine-tuning. For each pair it outputs one of entailment, neutral and contradiction by argmax over three classes. We run it twice, with the earlier answer as premise ($\ell_{\rightarrow}$) and with the later one ($\ell_{\leftarrow}$). Pairs with $\rho \ge 0.9$ are \textbf{copy}. Otherwise, the pair is \textbf{contradict} if either direction is contradiction, \textbf{restate} if both are entailment, \textbf{elaborate} if only $\ell_{\leftarrow}$ is entailment, \textbf{generalize} if only $\ell_{\rightarrow}$ is, and \textbf{shift} otherwise.


\subsection{Answer length} \label{subsec:length_check}

\begin{figure}[ht]
  \centering
  \begin{subfigure}{0.49\textwidth}
      \includegraphics[width=\linewidth]{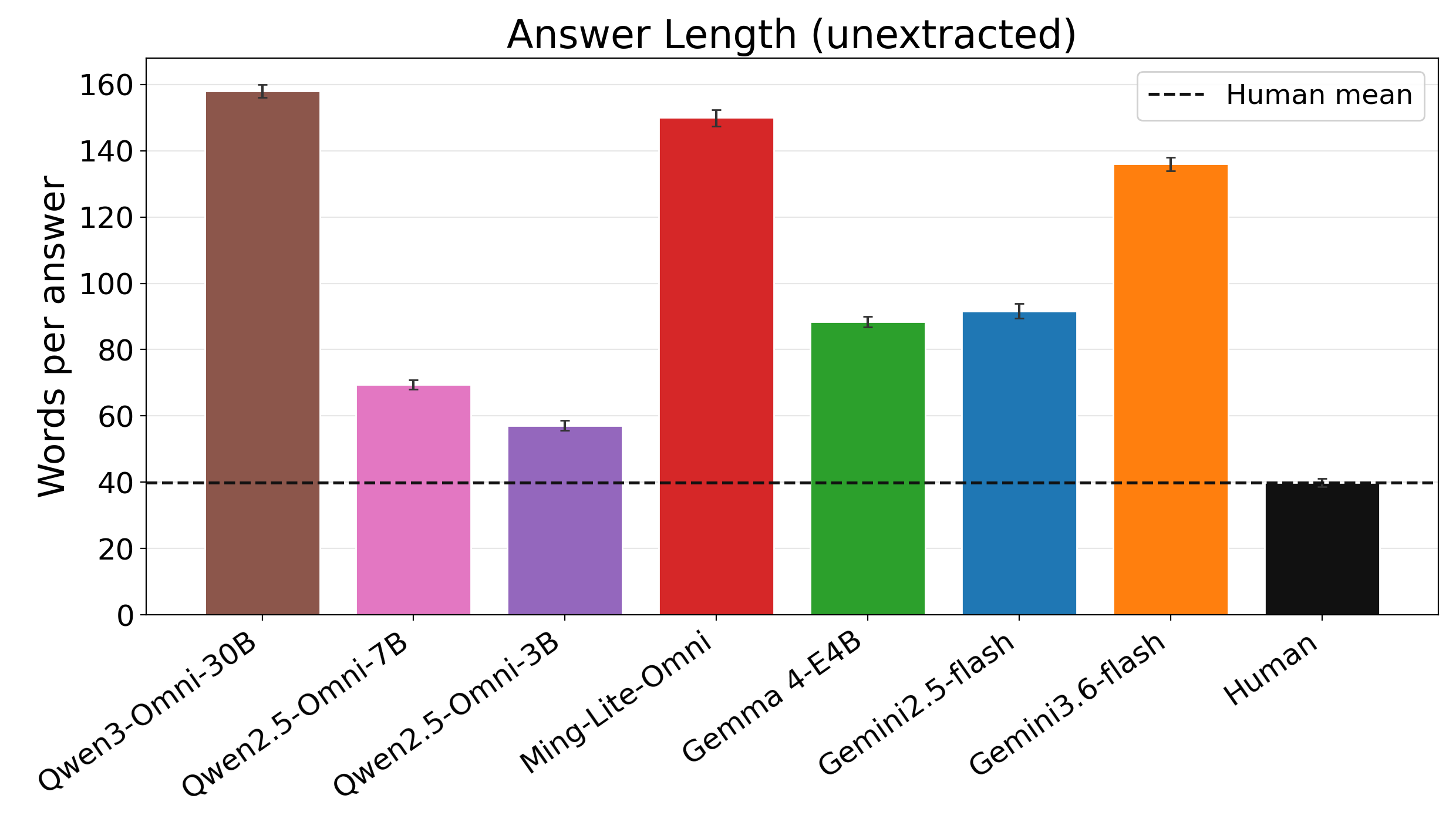}
      \label{fig:length_blob}
  \end{subfigure}
  \begin{subfigure}{0.49\textwidth}
      \includegraphics[width=\linewidth]{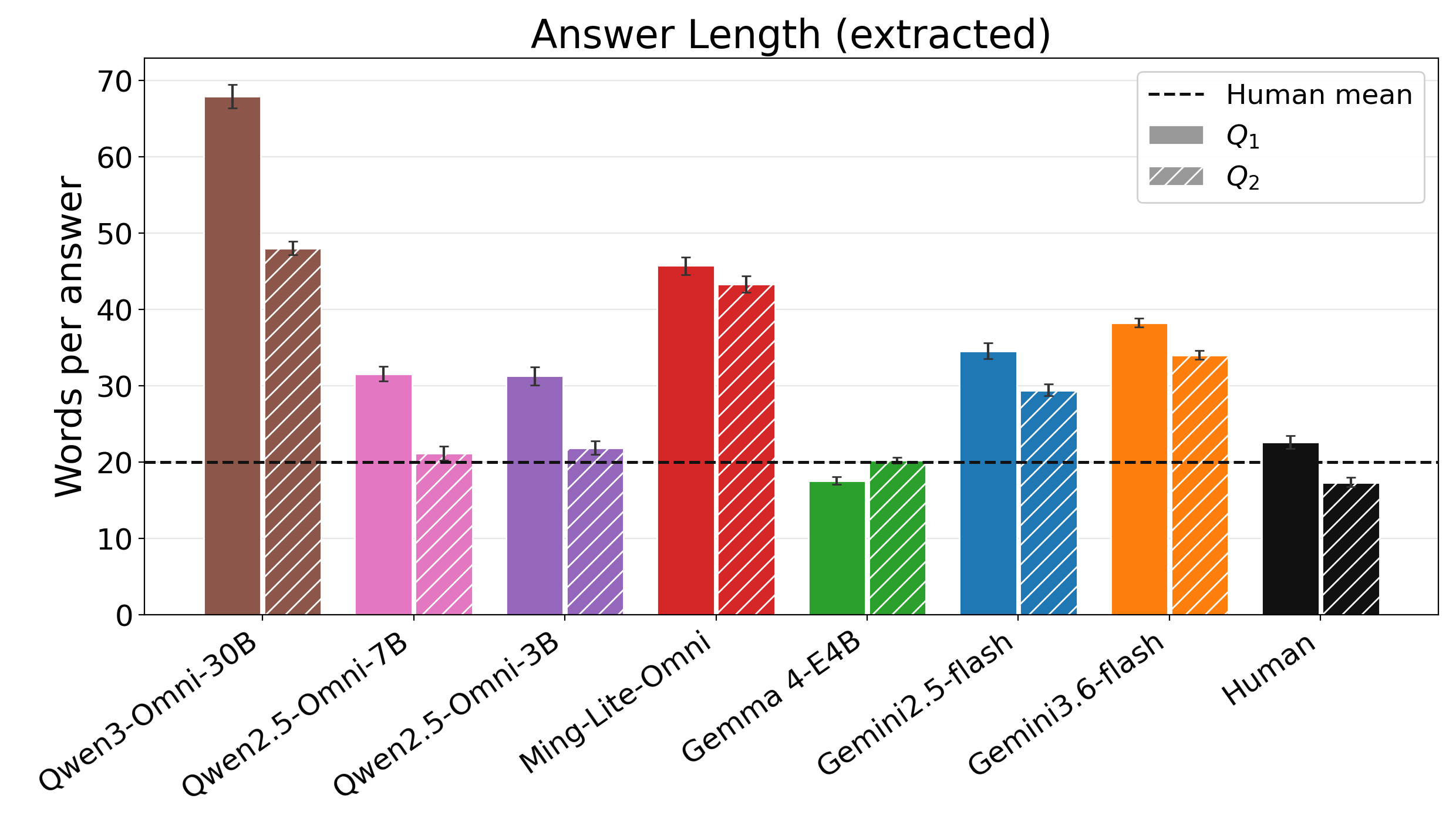}
      \label{fig:length_extracted}
  \end{subfigure}
  \caption{Comparison of answer lengths before and after extraction done on $\mathcal{T}_1$. The left part shows original answers while the right part shows after extraction. }
  \label{fig:length_check}
\end{figure}

\begin{figure}[ht]
  \centering
  \begin{subfigure}{0.49\textwidth}
      \includegraphics[width=\linewidth]{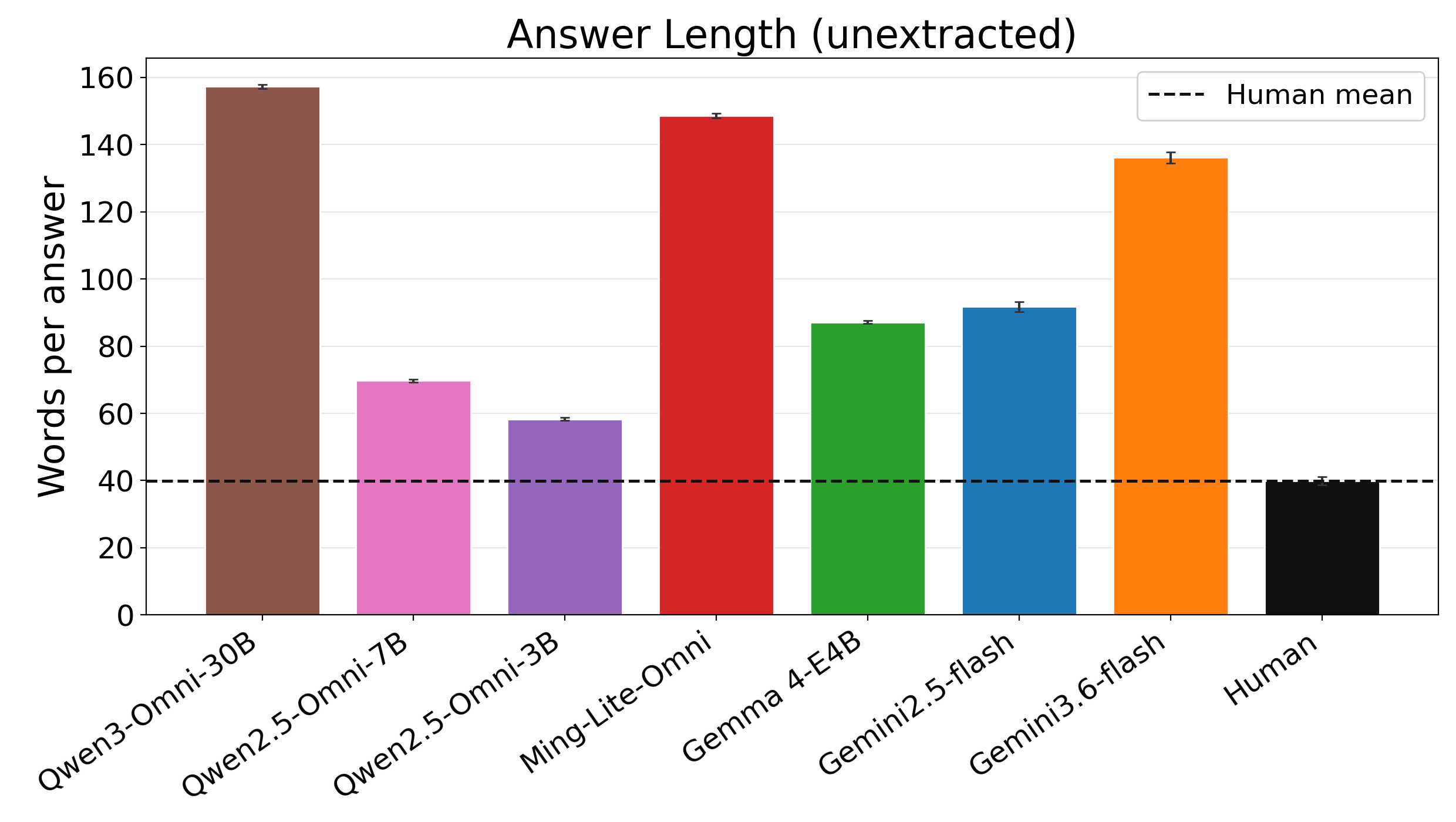}
      \label{fig:length_blob_a0}
  \end{subfigure}
  \begin{subfigure}{0.49\textwidth}
      \includegraphics[width=\linewidth]{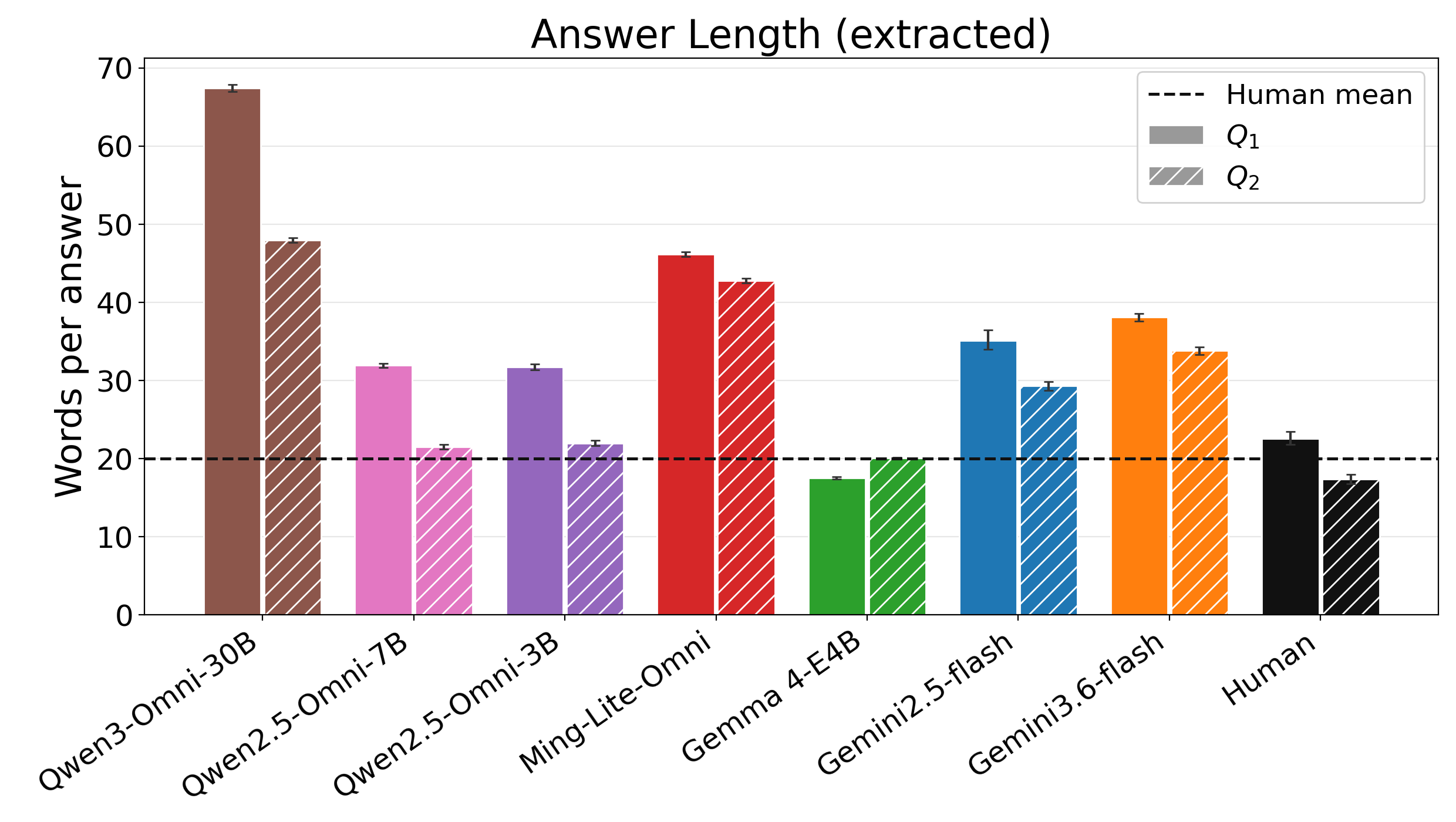}
      \label{fig:length_extracted_a0}
  \end{subfigure}
  \caption{Comparison of answer lengths before and after extraction done on \textbf{full set} when applicable. The left part shows original answers while the right part shows after extraction. }
  \label{fig:length_check_a0}
\end{figure}

To examing the loyalty of GPT extraction, also ablating on the possible influence of answer lengths on the RT metric, we report the distribution of answer length before and after extraction. Long, self-contained answers typically restate their own context and may entail their predecessors more often, which could lower a model's revision rate for reasons unrelated to perception. This check is done on $\mathcal{T}_1$. Figure~\ref{fig:length_check} shows words per extracted answer. Annotators answered around $20$ words on $Q_1$ and $18$ on $Q_2$, in comparable scale with most models except Qwen3-Omni-30B and Ming-Lite-Omni. Length also does not order the other models: Gemma and the Qwen2.5-Omni models don't write significantly longer answers than other perceivers, yet still change their readings sparingly(Figure~\ref{fig:revision}). This indicates that the difference observed in figure~\ref{fig:revision} is not a result of answer length or extraction but indeed related to the perception itself. In figure~\ref{fig:length_check_a0} we report the length check done on the \textbf{full set} with similar patterns.

\subsection{Value of updating} \label{subsec:rank_gain}
\begin{figure}[ht]
  \centering
  \begin{subfigure}{0.49\textwidth}
      \includegraphics[width=\linewidth]{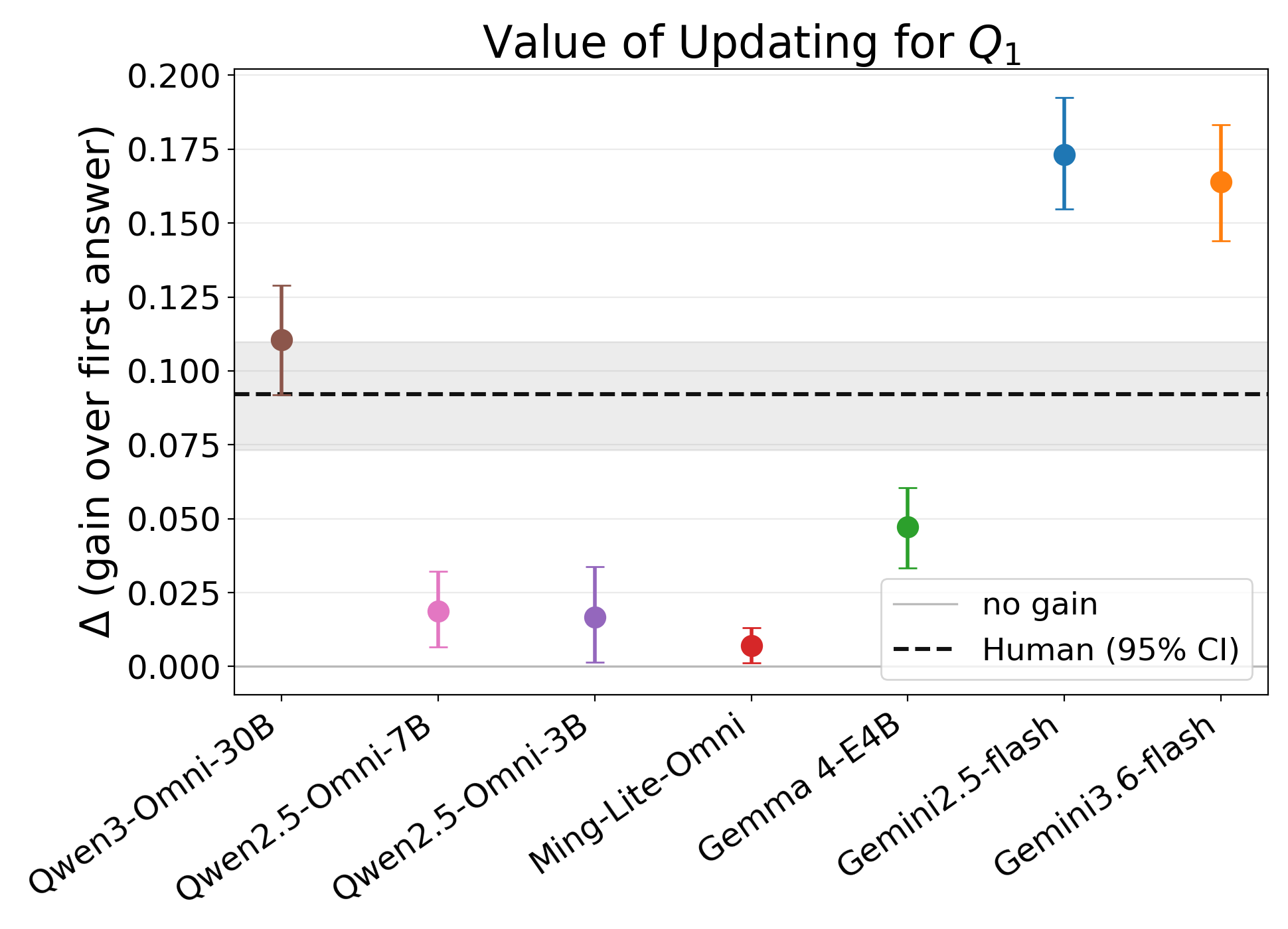}
      \label{fig:rank_gain_q1}
  \end{subfigure}
  \begin{subfigure}{0.49\textwidth}
      \includegraphics[width=\linewidth]{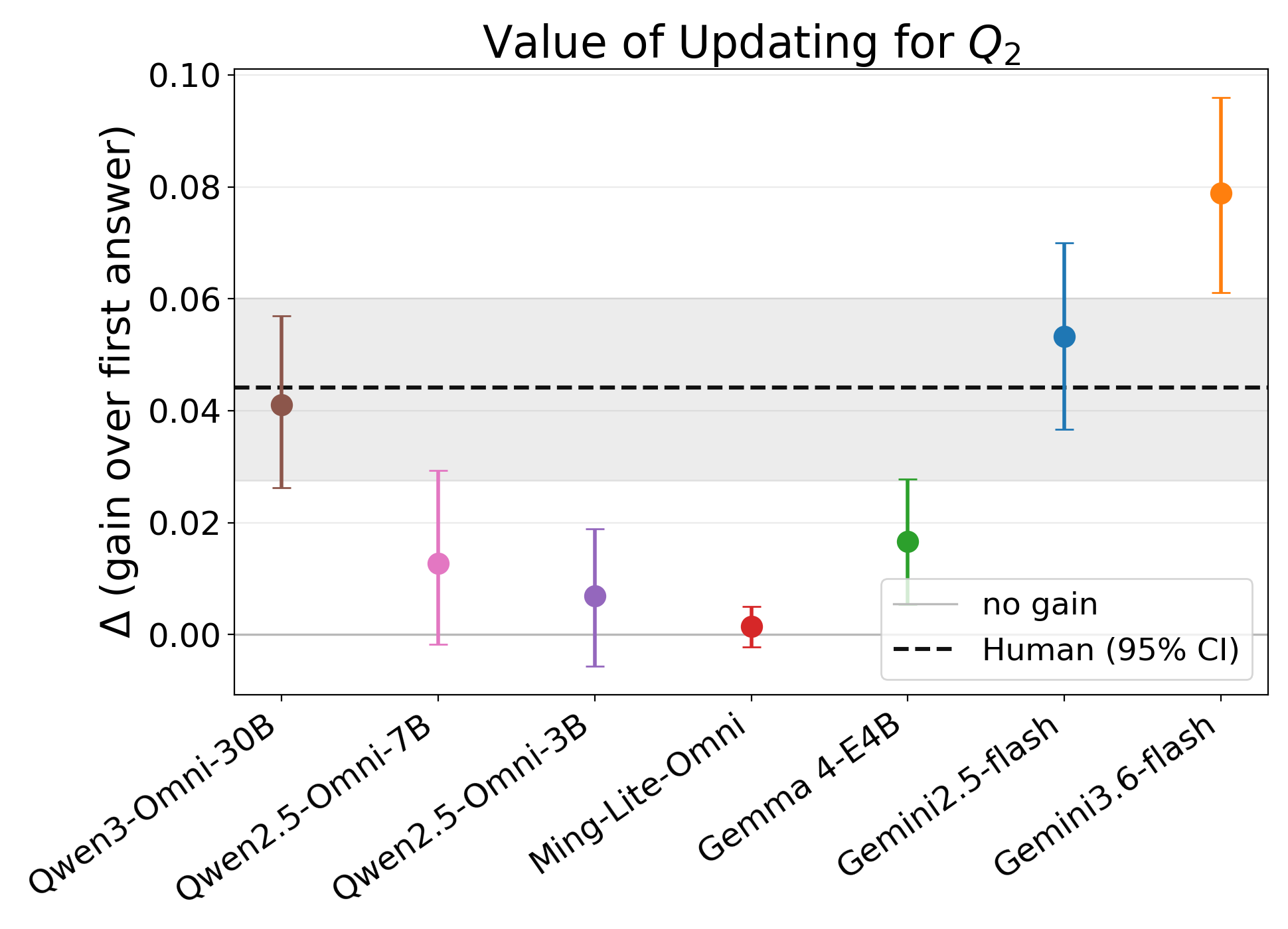}
      \label{fig:rank_gain_q2}
  \end{subfigure}
  \caption{Rank gain $\Delta$ computed for all models with human interval shown in black dotted line. Qwen2.5-Omni-7B/3B and Ming-Lite-Omni are close to 0, while Gemini models are gaining much more positively. The comparison should be made with $0$ instead of human standards.}
  \label{fig:rank_gain}
\end{figure}

Revising often is evidence of active perception only if the revisions help. We score each answer with the rank of Eq.~\ref{eq:rank} against what an annotator said \emph{from the same evidence} with target $H^a_Q\big(D_k(V)\big)$ and distractors $H^a_Q\big(D_{\kappa(k,V,V')}(V')\big)$ on other videos. Referring to the SCoRe's $\Delta(t_1,t_2)$~\citep{kumar2025training}, we then subtract the same rank computed with the perceiver's \emph{first} answer $A_Q\big(D_1(V)\big)$ in place of its current one:
\begin{equation}
  \Delta_Q(A,V,k) = \mathrm{R}\Big(A_Q\big(D_k(V)\big);\cdot\Big) - \mathrm{R}\Big(A_Q\big(D_1(V)\big);\cdot\Big).
\end{equation}
$\Delta$ is what the perceiver's updates added over repeating its first answer. A perceiver that copies forward scores exactly $0$, and $\Delta>0$ means later readings moved towards what people perceived.

We do this experimnet on $\mathcal{T}_2$ in alignment with the $\mathrm{Ant}$ measure. Figure~\ref{fig:rank_gain} shows that annotators gain around $+0.09$ on $Q_1$ and $+0.044$ on $Q_2$, so their revisions are productive rather than noise. Ming-Lite-Omni, Qwen2.5-Omni-7B/3B gain at most $0.021$ on either question, with intervals at or near zero: their whole cumulative sequence is worth little more than its first answer. The Gemini models gain more than annotators (above $+0.15$ on $Q_1$, around $+0.05$ and $+0.08$ for $Q_2$). This does not mean they update better than people: $\Delta$ is measured from each perceiver's own starting point, and annotators often start closer to the target (Figure~\ref{fig:specant}), leaving less headroom. We therefore read $\Delta$ against zero, not across perceivers.

\subsection{Gaze masking without audio} \label{subsec:gms_no_audio}


\begin{figure}[ht]
  \centering
    \begin{subfigure}{0.49\textwidth}
      \includegraphics[width=\linewidth]{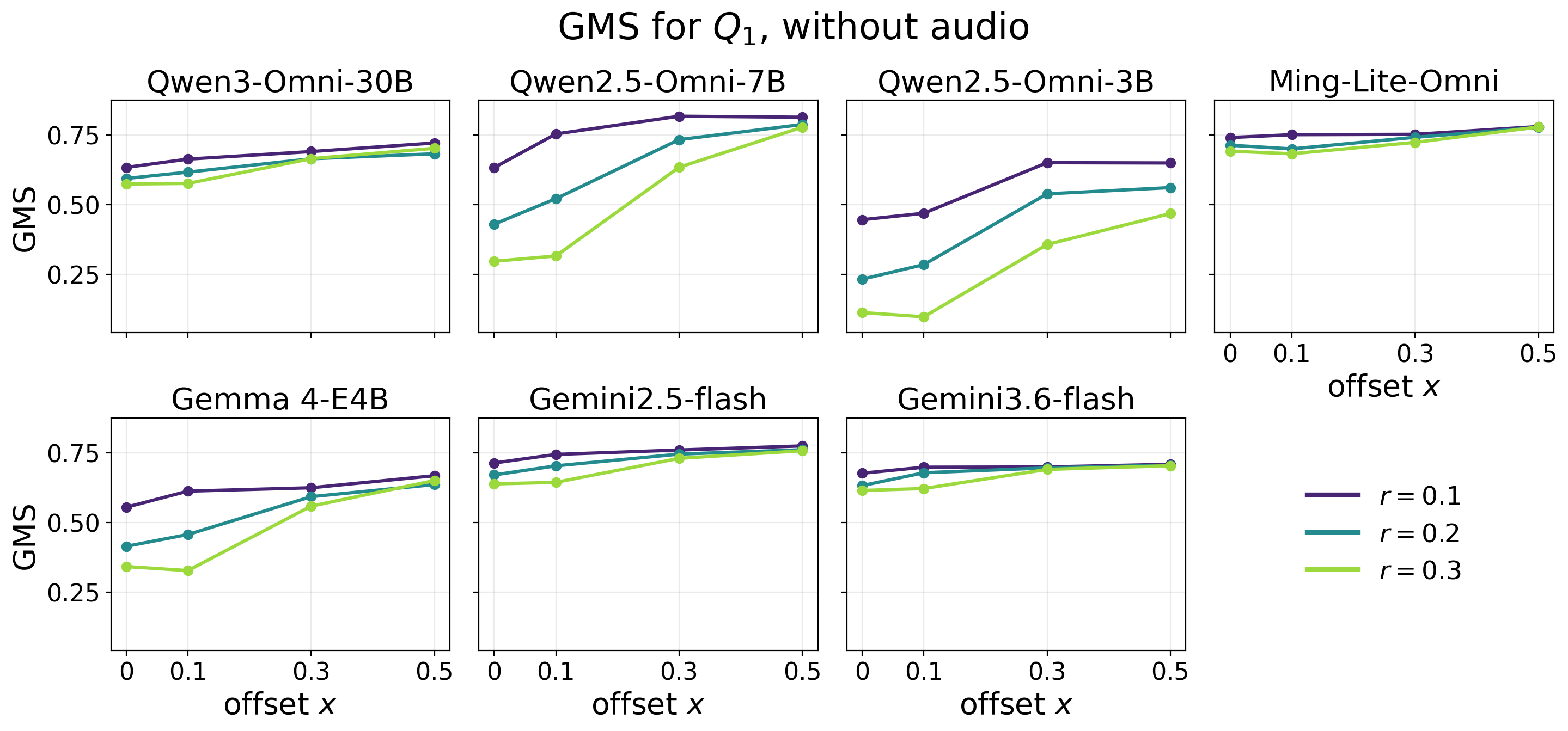}
      \caption{}
      \label{fig:gms_noaudio_q1}
  \end{subfigure}
    \begin{subfigure}{0.49\textwidth}
      \includegraphics[width=\linewidth]{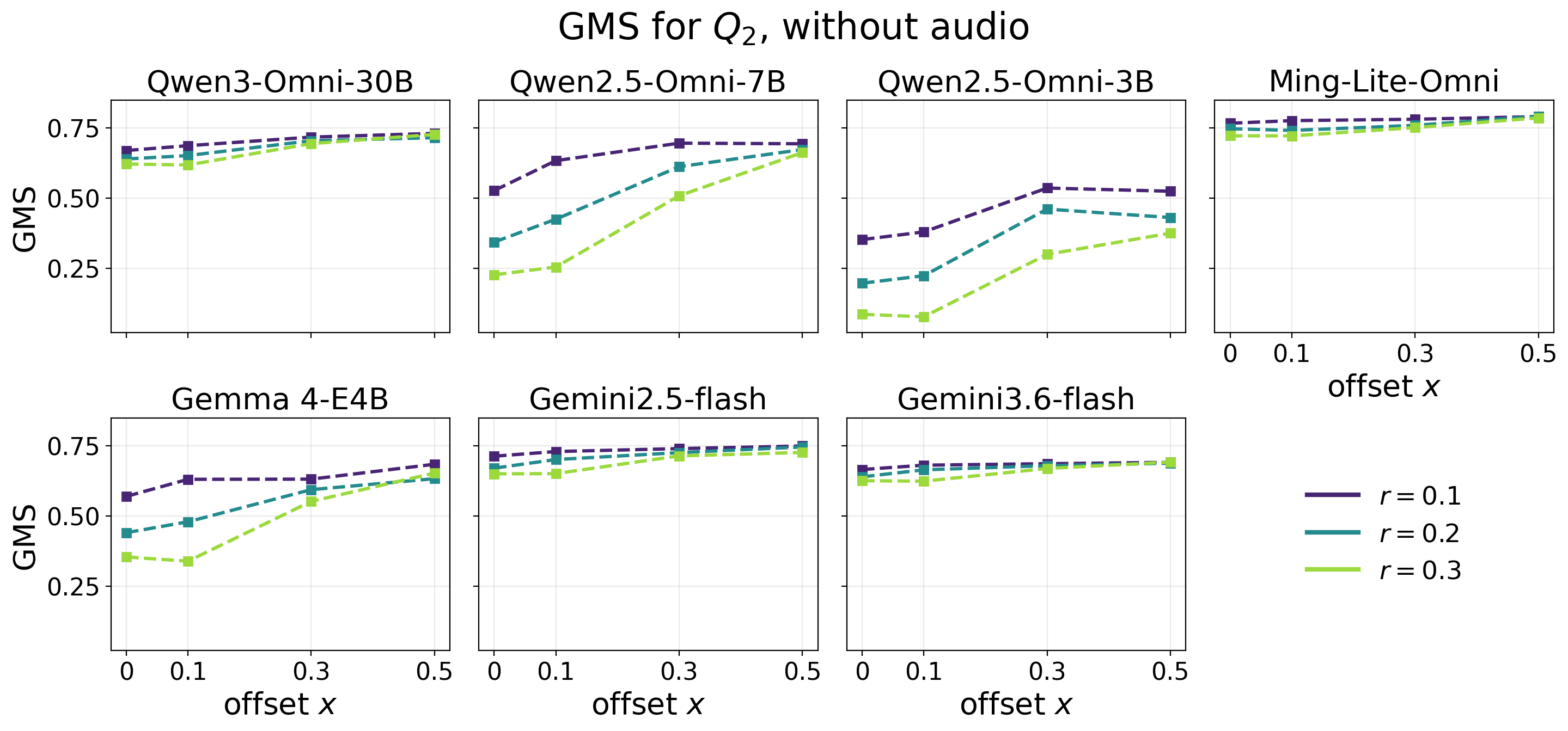}
      \caption{}
      \label{fig:gms_noaudio_q2}
  \end{subfigure}
  \caption{GMS plot for $Q_1$ and $Q_2$. Qwen2.5 models has lower GMS with larger $r$ and smaller $x$, showing they are sensitive to corruption. Gemma shows same pattern for $Q_1$ but not obvious for $Q_2$. Other models are not sensitive to gaze based manipulation. }
  \label{fig:gms_noduaio}
\end{figure}

Figure~\ref{fig:gms_noaudio_q1} and~\ref{fig:gms_noaudio_q2} compares the GMS without audio. We observe that without audio the effect of masking is significantly higher. For sensitive models such as Qwen2.5-Omni-7B/3B, the GMS drops below $0.20$ with $r=0.3$ and $x\leq 0.1$, indicating that they are even more influenced by the masking. On the other hand, Ming-Lite-Omni and Gemini models are still not being influenced sharply with varying $(x,r)$.

\end{document}